\documentclass[10pt,journal,compsoc]{IEEEtran}

\usepackage{booktabs} \usepackage{tabularx}
\usepackage[show]{notes-alt}
\newcommand{\mpara}[1]{\medskip\noindent{\bf #1}}
\usepackage[clock]{ifsym}
\usepackage{url} 
\usepackage{xcolor}
\usepackage{amsthm}
\usepackage{enumitem}
\usepackage{algorithm}
\usepackage{algorithmicx}
\usepackage[noend]{algpseudocode}

\usepackage{graphicx}
\usepackage{multirow}
\usepackage{booktabs}
\newcommand{\tabincell}[2]{\linespread{1.0}\selectfont\begin{tabular}{@{}#1@{}}#2\end{tabular}}
\usepackage{colortbl}
\usepackage{pifont}

\newcommand{\xmark}{\ding{55}}

\usepackage{amsmath,amssymb}
\usepackage{bbm}
\usepackage{caption,subcaption}
\usepackage{pdfpages}
\usepackage{footnote}
\usepackage{balance}
\usepackage{flushend}
\usepackage{color, colortbl}
\definecolor{Gray}{gray}{0.4}
\definecolor{lavendergray}{rgb}{0.77, 0.76, 0.82}
\definecolor{lightgray}{rgb}{0.83, 0.83, 0.83}
\definecolor{silver}{rgb}{0.75, 0.75, 0.75}
\definecolor{violet}{rgb}{0.74, 0.75, 0.75}
\definecolor{brown}{rgb}{0.74, 0.75, 0.75}
\definecolor{purple}{rgb}{0.74, 0.75, 0.78}

\newcommand{\deepwalk}{\textsc{DeepWalk}}

\newcommand{\lineemb}{\textsc{LINE}}
\newcommand{\verseemb}{\textsc{Verse}}
\newcommand{\nodetovec}{\textsc{Node2vec}}
\newcommand{\hope}{\textsc{HOPE}}
\newcommand{\dw}{\textsc{DeepWalk}}

\newcommand{\app}{\textsc{APP}}
\newcommand{\netmf}{\textsc{NetMF}}
\newcommand{\sdne}{\textsc{SDNE}}
\newcommand{\maxvote}{\textsc{Max-Vote}}
\newcommand{\graphsage}{\textsc{GraphSAGE}}
\newcommand{\graphsagegcn}{\textsc{GraphSAGE-GCN}}

\newcommand{\blog}{{{\textsf{BlogCatalog}}}}

\newcommand{\dblpa}{{{\textsf{DBLP-Au}}}}
\newcommand{\dblpc}{{{\textsf{DBLP-Ci}}}}

\newcommand{\cora}{{{\textsf{Cora}}}}
\newcommand{\youtube}{{{\textsf{Youtube}}}}
\newcommand{\reddit}{\textsf{Reddit}}
\newcommand{\epi}{{{\textsf{Epinion}}}}

\newcommand{\flickrl}{{{\textsf{Flickr}}}}
\newcommand{\twitter}{{{\textsf{Twitter}}}}
\newcommand{\pubmed}{{{\textsf{PubMed}}}}
\newcommand{\cocit}{{{\textsf{CoCit}}}}
\newcommand{\epinion}{{{\textsf{Epinion}}}}
\newtheorem{RQ}{RQ}
\newtheorem{prop}{Property}
\newtheorem{remark}{Remark}

\newtheorem{proposition}{Proposition}

\begin{document}

\title{A Comparative Study for Unsupervised Network Representation Learning}

\author{Megha~Khosla, Vinay~Setty, Avishek~Anand		\IEEEcompsocitemizethanks{
			\color{blue}{
			\IEEEcompsocthanksitem
			M. Khosla, and A. Anand are with the L3S Research center, Leibniz Universit\"at, Hannover, Germany.\protect\\
			E-mail: \{khosla,anand\}@L3S.de
			\IEEEcompsocthanksitem
			V. Setty is with the Department of Electrical Engineering and Computer Science, University of Stavanger,  Norway.\protect\\
			E-mail: vsetty@acm.org
			}
		}
	}
		
\maketitle

\begin{abstract}

There has been significant progress in unsupervised network representation learning (UNRL) approaches over graphs recently with flexible random-walk approaches, new optimization objectives and deep architectures. However, there is no common ground for systematic comparison of embeddings to understand their behavior for different graphs and tasks. 
We argue that most of the UNRL approaches either  model and exploit neighborhood or what we call context information of a node. These methods largely differ in their definitions and exploitation of context. Consequently, we propose a framework that casts a variety of approaches -- random walk based, matrix factorization and deep learning based -- into a unified context-based optimization function. We systematically group the methods based on their  similarities and differences. We study their differences which we later use to explain their performance differences (on downstream tasks). 

We conduct a large-scale empirical study considering 9 popular and recent UNRL techniques and 11 real-world datasets with varying structural properties and two common tasks -- node classification and link prediction. We find that for non-attributed graphs there is no single method that is a clear winner and that the choice of a suitable method is dictated by certain properties of the embedding methods, task and structural properties of the underlying graph. In addition we also report the common pitfalls in evaluation of UNRL methods and come up with suggestions for experimental design and interpretation of results.

\end{abstract}

\section{Introduction}

There has been a resurgence of unsupervised methods for network embeddings for graphs in the last five years~\cite{perozzi2014deepwalk,tsitsulin2018verse,ou2016asymmetric,NERD}. This is primarily due to improvements in modelling and optimization techniques using neural network based approaches, and their utility in a wide variety of prediction and social network analysis tasks such as link prediction~\cite{liben2007link}, vertex classification~\cite{grover2016node2vec}, recommendations \cite{ying2018graph},  knowledge-base completion etc.

\mpara{Lack of a comprehensive study.} In spite of their success, there is a lack of an in depth systematic study of the differences between various embedding approaches. 
Prior works have mainly focused on studying similarities between different embedding approaches using unifying theoretical frameworks~\cite{MF_WSDM2019,qiu2017network}. 
As we show in our experiments (cf. Section \ref{sec:results}), evaluation studies accompanying each new approach mostly focus on the experimental regimes where they perform well and omit the scenarios where they might perform sub-optimally. Comprehensive large-scale studies comparing these approaches under different experimental conditions are missing altogether to the best of our knowledge. 
Thus a fundamental practical question remains largely unanswered: From a comparative standpoint, \emph{Which Unsupervised Network Representation Learning (UNRL) approaches for nodes are most effective for different graph types and different downstream tasks?}

In this paper, we fill this gap by first proposing a common framework that focuses on the differences between the various UNRL approaches. 
Secondly, we perform a comprehensive experimental evaluation with 9  embedding methods (cf. Table~\ref{tab:algos-setup}) using different paradigms -- random walks, edge modeling, matrix factorization and deep learning -- that includes some of the earliest approaches for learning network representations to the more recent deep learning based approaches and 11 datasets (cf. Table~\ref{tbl:datasets}).
In this work we consider only unsupervised methods in the transductive scenario. 

\mpara{Unifying framework.} Our common framework for understanding UNRL approaches is inspired by the observation that most of the unsupervised learning approaches operate on an \emph{auxiliary neighborhood graph} in which similar vertices share an edge. We call such a  graph a \emph{context graph} and for any \emph{source} vertex $v$, its one-hop neighbor is called its \emph{context}. 
Depending on the embedding method, a vertex can be in the context of another if they are in the immediate neighborhood, reachable by  truncated random walks, or if they are in the same community/cluster etc. 
In this paper we study a wide variety of approaches-- e.g. that employ \emph{random walks}~\cite{perozzi2014deepwalk,grover2016node2vec,zhou2017scalable,tsitsulin2018verse}, \emph{neighborhood modelling}~\cite{tang2015line},  \emph{matrix factorization}~\cite{qiu2017network,ou2016asymmetric} and \emph{deep learning}~\cite{hamilton2017inductive,wang2016structural} -- in our unified framework, where these methods can be understood as optimizing a common form of objective function defined on their respective context graphs. 
This allows us to understand the differences among these approaches that arises due to their modelling of context.

\mpara{Comprehensive Experimental Evaluation.}
 In our evaluation of UNRL methods we investigate the \emph{conceptual differences between the embedding approaches that result in performance differences on downstream tasks.}  
 \textbf{First}, using graphs with diverse structural characteristics we argue about the utility of several approaches. We carefully chose 11 large network datasets (5 undirected and 6 directed) with diverse properties from social networks, citation networks, and collaboration networks.
With focus on reproducibility and large scale analysis, we chose at least one dataset used in each of the original papers and try to be as close to the authors original experimental setup as possible on four popular tasks -- node classification, link prediction, clustering and graph reconstruction. The first two tasks are thoroughly investigated in the main paper and all details corresponding to clustering and graph reconstruction tasks are only provided in the supplementary material. \textbf{Second}, in addition to performing a large-scale study with a large number of baselines we also find limitations in the experimental setup of earlier approaches. In particular, for evaluating link prediction performance in case of directed graphs most of the earlier works only check for the existence of an edge between a pair of nodes and ignore directionality of the edge. \textbf{Finally}, we question the claimed superiority of various embedding methods in the node classification task, wherein a na\"ive (yet effective) baseline is not considered. We surprisingly find that for several of the datasets comparable or even better performance is achieved by our improved na\"ive baseline.

\mpara{Key findings.} Our study does not propose a winner or a loser but highlights the strengths and weaknesses of approaches under different graph and task characteristics. We believe that our results can serve as guidelines for  researchers and industry practitioners in the choice of the wide range of embedding methods considered in this work. 
Some of the key findings of our study are as follows: 
\begin{itemize} 
\item Methods respecting the vertex's role as source and context during learning of representations as well as in their use for a task are recommended for link prediction in directed graphs.
\item Certain structural properties like clustering coefficient, transitivity, reciprocity etc. are recommended to be considered while choosing a specific method.
\item A simple immediate neighborhood based classifier turns out to be offering better or comparable performance for a number of datasets.

\end{itemize}

\section{Related Work}
With increasing number of unsupervised embedding methods, it has become extremely difficult to objectively compare and choose appropriate methods for a given dataset. Several existing surveys focus on categorization of various network embedding techniques with respect to the  methodology such as random walks, matrix factorization and edge modeling etc \cite{cai2018comprehensive, cui2018survey}.  But they fail to provide any unifying framework to compare and gain deeper understanding of various methods. Other works which do provide a common framework only focus on demonstrating the equivalence of various methods to matrix factorization \cite{MF_WSDM2019,qiu2017network,yang2015network} but do not consider the differences between the methods and their impact on task performance for a variety of graphs with different structural properties. Several other surveys~\cite{hamilton2017representationsurvey,wu2019comprehensive} consider a wide range of unsupervised and semi-supervised embedding methods without any empirical comparison. More importantly, the categorization and comparison in these surveys does not directly correspond with explaining why some methods are superior to other methods under certain circumstances. 
Surveys which include empirical comparison~\cite{goyal2018graph,zhang2018network} focus only on effect of training data size for various tasks or the effect of varying hyperparameters on task performance. In \cite{DBLP:journals/corr/abs-1811-05868}, the authors consider semi-supervised node classification and demonstrate the effect of different train/validation/test splits and hyperparameters on the performance of several graph neural network models.

We remark that the scope of this work includes unsupervised, transductive methods and non-attributed graphs. We include the most popular representative methods which follow our common objective function in our study. Other methods, for example based on generative modelling \cite{kipf2016variational}, adversarial training \cite{arga}, uncertainty modelling \cite{bojchevski2018deep} do not fall under our context based formulation and each of these categories deserves a separate study.
Semi-supervised methods like graph attention networks~\cite{velivckovic2017graph} are also not considered in our study.

\section{Unifying Framework and Research Questions}
\label{sec:theory}
In this section we first build a unifying framework in which we can conveniently cast the objective functions of the \emph{random walk}, \emph{matrix factorization} and \emph{deep learning} based Unsupervised Network Representation Learning (UNRL) methods. In particular, given a graph $G=(V,E)$ we are interested in learning low dimensional representations of each node $v\in V$ such that similar nodes in $V$ are embedded closer. These representations are then used for downstream tasks for example predicting missing links in $G$ or in node classification task where the goal is to predict missing node labels. Note that as we do not consider additional node or edge attributes in this work, the similarity information among the vertices is inferred from the topological structure of $G$.
Towards defining our common framework, we introduce the notion of a \emph{context graph} and propose a common objective function  into which all of the methods under investigation can be mapped.

\subsection{Context Graph}
\label{sec:contextgraph} 
 In order to understand how various methods differ in their definitions and treatment of similarity, we begin by constructing an auxiliary \emph{directed} and \emph{weighted} graph $\mathcal{C}$ from $G$ where $\mathcal{C}=(V,E')$ such that the higher the weight of edge $(u,v)\in E'$ the larger the similarity among nodes $u$ and $v$. Moreover for nodes with no edge between them, we construct an edge with weight $-1$. The negative weight here denotes the dissimilarity between two nodes.
 We call $\mathcal{C}$ as the \emph{context graph} and for each edge $(u,v) \in E'$, $v$ is called the \emph{context} of node $u$. Let $C$ denote the corresponding adjacency matrix of $\mathcal{C}$ with $c_{i,j}$ denoting $(i,j)th$ element in $C$. For an edge $(u,v)\in E'$, we call $u$ as the \emph{source} node and $v$ its \emph{context}.

As each node can be a source or a context of some other node, we denote the  \emph{source} and \emph{context} representations of nodes in $\mathcal{C}$ by $\Phi \in \mathbb{R}^{|V|} \times \mathbb{R}^d $ and $\theta \in \mathbb{R}^{|V|} \times \mathbb{R}^d $ respectively. For any node $v_i$, $\Phi_i$ and $\theta_i$ represent respectively its d-dimensional source and context vectors (representations).
We are then interested in learning  $\Phi$ and $\theta$ while minimizing the following loss
\begin{equation} 
\label{eq:loss}
\mathcal{J} =  -\sum_{i,j} c_{i,j}\cdot f(\Phi_i,\theta_j) ,  
\end{equation}
where, $f$ is monotonically increasing in $\Phi_i \cdot \theta_j$.

We recall that by our construction, for any two dissimilar nodes $i,j$, $c_{i,j}=-1 $. This imposes an additional constraint on the embedding vectors such that the corresponding dot product of embedding vectors is minimized for dissimilar nodes.
Note that by minimizing \eqref{eq:loss} a vertex (in its source representation) will be embedded closer to its context (in its context representation); and therefore two vertices sharing same context will be embedded closer (in their source representations) by transitivity. 
In this work we consider \deepwalk~\cite{perozzi2014deepwalk}, \nodetovec~\cite{grover2016node2vec}, \app~\cite{zhou2017scalable}, \lineemb-2~\cite{tang2015line} employing the above form of the loss function. We also include two methods which use matrix factorization objectives--\hope~\cite{ou2016asymmetric} and \netmf~\cite{qiu2017network}, based on their equivalence to the above objective demonstrated in \cite{MF_WSDM2019}.

In some works such as \verseemb~\cite{tsitsulin2018verse}, \lineemb-1~\cite{tang2015line}, \sdne~\cite{wang2016structural}, unsupervised \graphsage~\cite{hamilton2017inductive} the loss function does not consider separately the context role of a vertex. More specifically, for any two neighboring nodes they attempt to embed vertices closer by maximizing the dot product of their source representations. Formally, they learn only $\Phi$ by minimizing the following loss function. 
\begin{equation} 
\label{eq:loss2}
\mathcal{J'} =  -\sum_{i,j}c_{i,j}\cdot f(\Phi_i,\Phi_j).
\end{equation}

While many of the considered methods can be explained under a unified framework based on their similarities in their objective functions (as also done by previous works \cite{qiu2017network, MF_WSDM2019}), we are interested in understanding their differences due to their modelling decisions.
\begin{table}[h!]
	\centering
		\scalebox{0.95}{
			\begin{tabular}{|c|c|}
				\hline
				  \bf \tabincell{c}{Symbol/ \\Text} & \bf Meaning \\
				\hline
			$G$ & \tabincell{c}{Undirected or directed graph with \\ vertex set $V$ and edge set $E$} \\
			\hline
			$\mathbf{P} =D^{-1}A$  & \tabincell{c}{Transition matrix where A and D \\ are the adjacency and degree matrices of $G$} \\
			\hline
			$d_i$ & out-degree of a vertex $v_i$ in $G$
			\\
						\hline
			$\mathcal{C}$ & \tabincell{c}{Context graph of $G$} \\
			\hline
			$C$ & \tabincell{c}{Adjacency matrix of $\mathcal{C}$} \\
			\hline
			context & \tabincell{c}{One-hop neighbor of a vertex in $\mathcal{C}$}\\
			\hline
			$\Phi,\theta$ & Source and Context representation matrices \\
			\hline
		\end{tabular}}
		\caption{Notations}
		\label{tab:notations}
	\end{table}

\begin{table*}[ht!!!]
		\tabcolsep 5pt

		\centering
		
		\scalebox{1}{
			\begin{tabular}{|c|c|c|c|c|c|c|}
				\hline
				  \bf Context Graph&\bf Algorithm &\bf \tabincell{c}{Sym. C } & \bf \tabincell{c}{Learnt \\ Embeddings} & \bf \tabincell{c}{Used \\ Embeddings}& \bf Loss &\bf Optimization \\
				\hline
		Random Walk based&	\dw & \checkmark & $\Phi, \theta$  &$\Phi$ &$ -\sum_{i,j}c_{i,j} \log {\exp{(\Phi_i \cdot \theta_j)} \over \sum_{k\in V} \exp(\Phi_i \cdot \theta_k) } $ & \tabincell{c}{Hierarchical   Softmax }\\
		\hline
				
	Random Walk based &	 \nodetovec & \checkmark  & $\Phi, \theta$  &$\Phi$ & $-\sum_{i,j}c_{i,j} \log {\exp{(\Phi_i \cdot \theta_j)} \over \sum_{k\in V} \exp(\Phi_i \cdot \theta_k) } $ &\tabincell{c}{Negative Sampling (NS) } \\
			\hline
			
       PPR based & \app & \xmark  & $\Phi, \theta$  & $\Phi, \theta$& $-\sum_{i,j}c_{i,j} \log {\exp{(\Phi_i \cdot \theta_j)} \over \sum_{k\in V} \exp(\Phi_i \cdot \theta_k) } $& NS\\
        \hline
        PPR based &   \verseemb  &\xmark  &$\Phi$  &$\Phi$ & $-\sum_{i,j}c_{i,j} \log {\exp{(\Phi_i \cdot \Phi_j)} \over \sum_{k\in V} \exp(\Phi_i \cdot \Phi_k) } $& NS  \\
        \hline
        Adjacency based & \lineemb-1 & \checkmark & $\Phi$  &$\Phi$ & $-\sum_{i,j}c_{i,j} \log {1 \over 1 + \exp(-\Phi_i \cdot \Phi_j) } $ & NS\\
         \hline   
        Adjacency based&  \lineemb-2 & \xmark & $\Phi, \theta$  &$\Phi$ & $-\sum_{i,j}c_{i,j} \log {\exp{(\Phi_i \cdot \theta_j)} \over \sum_{k\in V} \exp(\Phi_i \cdot \theta_k) } $& NS\\
          \hline
       Direct matrix &   \netmf  &\xmark  & $\Phi, \theta$ & $\Phi $& $|| C - \Phi\cdot \theta||^2_F$ &\tabincell{c}{Matrix  Factorization (MF) }\\
          \hline
        Direct matrix &  \hope &\xmark  & $\Phi, \theta$ & $\Phi, \theta$ &$|| C - \Phi\cdot \theta||^2_F$ & MF\\
\hline
 Adjacency based&\sdne & \xmark  & $\Phi$ & $\Phi$ & see Equation~\eqref{eq:sdne}&\tabincell{c}{Deep  Autoencoders }\\
 \hline
Random Walk based &\tabincell{c}{ Unsupervised \\ \graphsage} & \checkmark & $\Phi$ &$\Phi$  &$-\sum_{i,j}c_{i,j} \log {\exp{(\Phi_i \cdot \Phi_j)} \over \sum_{k\in V} \exp(\Phi_i \cdot \Phi_k) } $ & \tabincell{c}{NS with \\ Neighborhood Aggregation } \\
 \hline
		\end{tabular}}
		\caption{\small{A summary of \emph{Network Representation Learning} algorithms with respect to Context and Optimization. \textbf{Sym.C} corresponds to if the adjacency matrix of context graph is symmetric .}}
		\label{tab:algos-setup}
	\end{table*}
 In the rest of this section, we elaborate these differences based on (1) how they \emph{define context} or neighborhood in the corresponding context graph, (2) how they \emph{exploit context} and (3) how they \emph{optimize their objectives}. In Section~\ref{sec:cg} we list 4 different schemes of defining context with each scheme having examples of two embedding methods. For these chosen methods, we then focus on exploitation of \emph{context} in Section~\ref{sec:contextexploitation}. We then elaborate on various optimization approaches used by these methods in Section~\ref{sec:opt}. The main discussed differences and similarities among the studied methods is also summarized in Table~\ref{tab:algos-setup}. Finally, in Section~\ref{sec:rq}, we formulate a set of research questions which are answered based on the differences elaborated below and the experimental results in Section~\ref{sec:results}. The main notations used in the rest of the paper are listed in Table~\ref{tab:notations}.

\subsection{Different Schemes of Defining Context}
\label{sec:contextschemes}
    In this section, we compare various approaches with respect to the different ways in which the context graph is defined in Section \ref{sec:contextgraph}. The simplest possible context matrix which can be used is the adjacency matrix itself, i.e., nodes sharing a link are similar to each other. In this case the context graph is the same as the original graph $G$. While some methods directly use similarity notions like Katz similarity \cite{katz1953new}~(e.g., \hope) or Personalized PageRank (e.g., \verseemb) to quantify similarity among vertices, other methods explore higher order neighborhoods via random walks and  quantify similarity among nodes by their co-occurrence in these walks (e.g., \deepwalk~and \nodetovec).
\label{sec:cg}
\subsubsection{Random Walk Based Context}
In random walk based methods, the higher order neighborhoods are usually sampled to define the context graph. Roughly, a vertex pair $(u,v)$ co-occurring in a random walk will correspond to two directed edges in the context graph : $u\rightarrow v$ and $v\rightarrow u$. In the first edge $v$ serves as a context for $u$ while for the second edge $u$ is the context. We explain below more precisely the context graphs of two popular methods under this category, namely \deepwalk~and \nodetovec.  

\mpara{\deepwalk, \nodetovec{} .}  These methods employ truncated random walks of length $T$ from each vertex $v\in V $ to create vertex sequence, say $W_v$. In particular, for each $v_i \in W_v$ and for each $v_j \in W_v[i-r:i+r]$ (r is the window size), $(v_i,v_j)$ forms an edge in the corresponding context graph.
While \deepwalk{} performs a uniform random walk, \nodetovec{} follows a 2nd order random walk.

More specifically, from \cite{qiu2017network}, for any pair of vertices $v_i,v_j \in V$ for \deepwalk's walk lengths $T$ and window size $r$ we have 
 \begin{equation}
 \label{eq:deepwalk}
  {c_{i,j}\over \sum_{u,v| c_{u,v}>0} c_{u,v}} \xrightarrow{p}{1\over 2T}\sum_{r=1}^T \left( {d_i\over \sum_{i}d_i}\cdot (\mathbf{P}^r)_{i,j} + {d_j\over\sum_{i}d_i}\cdot (\mathbf{P}^r)_{j,i}\right),
  \end{equation}
Note that the ratio in the L.H.S of \eqref{eq:deepwalk}, i.e. the ratio of weight of edge $(v_i,v_j)$ to the total (positive) edge weight in the context graph, quantifies the similarity of vertices $v_i$ and $v_j$.
The similarity between $v_i$ and $v_j$ is therefore proportional to the sum of probabilities that vertex $v_j$ is reachable from a $r$-truncated random walk started from the vertex $v_i$ and vice versa. Here $r$ varies from $1$ till the original walk length $T$. 
One of the implications of \eqref{eq:deepwalk} is that $c_{i,j}$ and $c_{j,i}$ both represent exactly the same quantity as the r.h.s  is symmetric in $i$ and $j$. In general the probability of reaching $v_i$ to $v_j$ by a random walk of $r$ hops i.e. $(\mathbf{P}^r)_{i,j}$ is not the same as the probability of reaching from $v_j$ to $v_i$ in $r$ hops ($(\mathbf{P}^r)_{j,i}$), the presence of both the terms in r.h.s make the similarity between source $v_i$ and context $v_j$ equal to similarity between source $v_j$ and context $v_i$. Therefore, even if the context is explicitly represented by learning a context representation for each node, the possible asymmetric properties of source and context are completely  ignored. For example, consider two vertices in a directed graph or in an undirected graph with very different local neighborhood structures such that the reachability probability of one node to another is not the same in the other direction.

Considering  the biased walks in \nodetovec{}, it first computes a second order transition probability to sample the next vertex in the walk as defined below.
\[ 
\mathbf{\underline{P}}_{u, v, w} = {T_{u\rightarrow v\rightarrow w} \over \sum_w T_{u\rightarrow v\rightarrow w}, }
\]
where
\begin{equation}
\label{eq:transition}
T_{u\rightarrow v\rightarrow w} = \begin{cases} {1 \over p}, \text{if } (u,v)\in E (v,w)\in E, u=w  \\
        {1 } ,\text{if } (u,v)\in E (v,w)\in E, u\neq w, (u,w)\in E\\
        {1\over q } ,\text{if } (u,v)\in E (v,w)\in E, u\neq w, (u,w)\not\in E\\
		0, \text{otherwise}
	\end{cases}. \end{equation}
From \eqref{eq:transition}, we note that for directed graphs where an edge $(u,v)$ does not automatically apply existence of an edge $(v,u)$, $p$ might have limited or no influence over the random walks, for example for directed graphs with zero reciprocity\footnote{In a directed network, the reciprocity equals to the proportion of edges for which an edge in the opposite direction exists}. Again for graphs with low clustering coefficient, condition 2 might not hold for many cases. In summary, biased walks of \nodetovec{} might be reduced to uniform random walks as employed by \deepwalk{} for graphs with low reciprocity and low clustering coefficient and transitivity.

Under assumptions of infinite length walks on undirected graphs, it can be shown that for \nodetovec{}
 \begin{equation}
 \label{eq:node2vec}
 {c_{i,j}\over \sum_{u,v|c_{u,v}>0} c_{u,v}} \xrightarrow{p}{1\over 2T}\sum_{r=1}^T\sum_k \left( {\mathbf{X}_{k,i}} (\mathbf{\underline{P}}^r)_{k,i,j} + {\mathbf{X}_{k,j}} (\mathbf{\underline{P}}^r)_{j,k,i}\right),
  \end{equation}
where $\mathbf{X}$ represents a stationary distribution of the second order random walk. Comparing it with \eqref{eq:deepwalk}, uniform degree distribution to choose the source vertex and the transition probabilities are respectively replaced with stationary distribution of the second order random walk and the second order transition probability.
As  \eqref{eq:node2vec} is again symmetric in $i,j$, we also obtain a symmetric adjacency matrix for context graphs used by \nodetovec{}.
In summary, the respective context graphs have the following properties.

  \begin{prop}
  \label{prop:symm}
  For \deepwalk{} and \nodetovec{}'s context graphs, the roles of a vertex as source and context are indistinguishable by construction, for example $c_{i,j}$ and $c_{j,i}$ will be identical even if the underlying graph $G$ is directed.
  \end{prop}
     \begin{prop}
     \label{prop:param1node2vec}
     The parameter $p$ as used by \nodetovec{} will have no effect for a directed graph with zero reciprocity as there are no back edges.
     \end{prop}
    \begin{prop} 
    \label{prop:param2node2vec}
    For any triplets $u,v,w$ we have $\mathbf{P}_{v,w} = \mathbf{P}_{u,v,w}$ when $u,v,w$ do not form a triangle and there are no back edges, i.e., $u\neq w$. This implies that \nodetovec{}'s biased walks might not give any additional advantage in case of graphs with low transitivity, low clustering coefficient and zero reciprocity.
\end{prop}

\subsubsection{Personalized PageRank (PPR) Based Context}
\mpara{\app{} and \verseemb.} These methods use random walks with restarts to draw source-context pairs. In particular, every time a  walk starts from a vertex chosen uniformly at random; a walk is continued with probability $1-\alpha$ where $\alpha$ is the predefined restart probability. The first and last vertices of this walk forms a directed edge in the corresponding context graph with the first node of the walk being the source node. For any vertices $i,j \in V$ we compute the theoretical estimate of  $c_{i,j} \over \sum_{u,v} c_{u,v} $.
 \begin{proposition}
 \label{prop:app}
 Let $i\in V$ be uniformly chosen as done in \app{}.
 Let $j\in V$ be such that the shortest hop distance between nodes $i$ and $j$ be $h$.
Then node $j$ is the last node in the walk starting from $i$ with probability $O({1 \over |V|}(1-\alpha)^h\cdot \alpha\cdot (\mathbf{P}^h)_{i,j})$, i.e.,
 
 $$ {c_{i,j} \over \sum_{u,v|c_{u,v}>0} c_{u,v} } = O\left({1 \over |V|}(1-\alpha)^h\cdot \alpha\cdot (\mathbf{P}^h)_{i,j}\right).
 $$
 
\end{proposition}
 The proof for above proposition is provided in the supplementary material. We observe that the context graph used by \app{} and \verseemb{} is considerably different from that used by \deepwalk{} and \nodetovec{}. Note Equations \eqref{eq:deepwalk} and \eqref{eq:node2vec} imply $c_{i,j}=c_{j,i}$ for \deepwalk{} and \nodetovec{} whereas this is not the case for \app{} and \verseemb{}, where their values depend on the neighborhood structures of nodes as shown in Figure~\ref{fig:app} where there is a higher probability of reaching from node $u$ to $v$ than vice-versa. We therefore have the following property.
 \begin{prop}
 \label{prop:app}
 For any $i,j \in V$, $c_{i,j}$ is not always equal to $c_{j,i}$, i.e., $C$ is not always a symmetric matrix or the similarity relation between vertices is not always symmetric.
 \end{prop}

\subsubsection{Adjacency Based Context}
 \mpara{\lineemb~ and \sdne.}  These methods 
 directly use the given graph as its context graph, i.e., $C=A$.
 They aim to embed vertices closer which have either links between them (optimizing for first order proximity) or share common 1-hop neighborhood (optimizing for second order proximity). They specifically differ in their exact formulations of loss functions and optimization strategies which will be discussed in detail in Section~\ref{sec:opt}. Corresponding to \lineemb{}, we study both of its variants : \lineemb-1  (  optimizing only  first-order proximity) and \lineemb-2  (optimizing  only second-order  proximity). \lineemb-1+2  is  obtained  by  normalizing  and  concatenating  the  embedding vectors from \lineemb-1 and \lineemb-2
 
 \mpara{Special Case of Unsupervised \graphsage{}.} \graphsage{} uses a two layer deep neural architecture where in each layer $k$ a node $v\in V$ computes its representation $h^k$ as an aggregation of representations (from previous layer) of its neighbors,  $\{h^{(k-1)}(u),\forall u \in N(v)\}$. The parameters of aggregation functions are learnt using the loss function similar to \deepwalk{}. In other words, \graphsage{} also optimizes for embedding the vertices closer, which are more similar with respect to the context matrix generated using Equation \eqref{eq:deepwalk}, where, an source embedding vector of  a vertex is a function of embedding vectors of its immediate neighbors.
 Intuitively this implies that vertices having links between them will be embedded closer.
For \graphsage~we report the best results corresponding to one of its four aggregators (Mean, MeanPool, MaxPool and LSTM). In addition, we  study GCN variant of \graphsage{} where the aggregator function is the graph convolution network. Note that we used the unsupervised and transductive variant of \graphsage{} for this work.

 \subsubsection{Direct Matrix Based Context}
  \label{sec:context}
\mpara{\netmf.} \netmf{} is derived from a theoretical analysis of \deepwalk{} and directly factorizes  the context matrix with $(i,j)$th element given by
\begin{equation}
\label{eq:netmf}
    c_{i,j} = \log \left( {vol(G)\over kT} \left({1\over d_j} \cdot{(\sum_{r=1}^T \mathbf{P}^r})_{i,j} \right)  \right)
\end{equation} 
where $T,r,k$ are hyperparameters and correspond to walk length, window size and negative samples in \deepwalk.  

We remark here that while \deepwalk{} explicitly encodes similarity between vertices as given by Equation \eqref{eq:deepwalk}, using the equivalence of SGNS\cite{mikolov2013distributed} optimization to matrix factorization, \cite{qiu2017network} proposes that \deepwalk{} implicitly factorizes the context matrix with its $(i,j)th$ element given by Equation \eqref{eq:netmf}. Note that the focus of this work is not to validate/invalidate this connection but understand the kind of vertex similarities different methods try to encode in its latent representations. \deepwalk{}  and \netmf{} are therefore not only different from their optimization techniques but also their respective context graphs representing similarities between vertices.

\mpara{\hope.} This approach preserves the asymmetric role information of the nodes by approximating high-order proximity measures like Katz measure \cite{katz1953new}, Rooted PageRank \cite{song2009scalable} etc. We study the version of \hope~where it uses Katz similarity matrix as the context matrix as it also gives us a different type of context graph to compare with. For example, the context graph generated for Rooted PageRank is quite similar to the ones used by \verseemb{} and \app{}. In a Katz similarity matrix, each entry $c_{i,j}$ is a weighted summation over the path set between two vertexes. More specifically,
$ c_{i,j} = \sum_{\ell=1}^\infty \beta^k (A^k)_{i,j},$
where $\beta$ is a decay parameter and determines how fast the weight of the path decay with growing length.

\subsection{Exploitation of Context}
\label{sec:contextexploitation}
 Methods differ in their learning and usage of context representations. While some methods only learn source representation for a node, other methods learn both representations but only utilize source representations for the downstream task. There is yet another class of methods which in addition to learning two representations also use both of them for downstream tasks.

\mpara{\deepwalk{}, \netmf{}, \lineemb{}-2 and  \nodetovec{}.}
\label{sec:rolecontext}
These methods learn both source and context representations but use only source representation for the downstream tasks.

\mpara{\graphsage{}, \lineemb{}-1, \verseemb{} and \sdne.} All of these methods learn only a source representation of a  vertex and ignore its representation as a context.

\mpara{\app{} and \hope{}.} Both these methods learn two representations per vertex and use both the representations for downstream tasks. They in fact use the context representation to represent the node in its destination role if the original graph $G$ is directed.

\mpara{Difference between \app{} and \verseemb{}.} \app{} and \verseemb{} both perform random walks with restarts to compute their respective context graphs. As already discussed (cf. Property \ref{prop:app}), the similarities encoded by the context graph in their case are not symmetric, yet \verseemb{} ignores this asymmetries and attempts to encode the similarities between vertices in  a single embedding space. This is quite contrary to its motivation of encoding Pensonalized PageRank (PPR) which is by construction asymmetric, i.e., $PPR(i,j)$ is not always equal to $PPR(j,i)$, where $PPR(i,j)$ represents the PPR of $i$ with respect to $j$ .

\begin{figure}
\begin{center}
\includegraphics[width =0.17\textwidth]{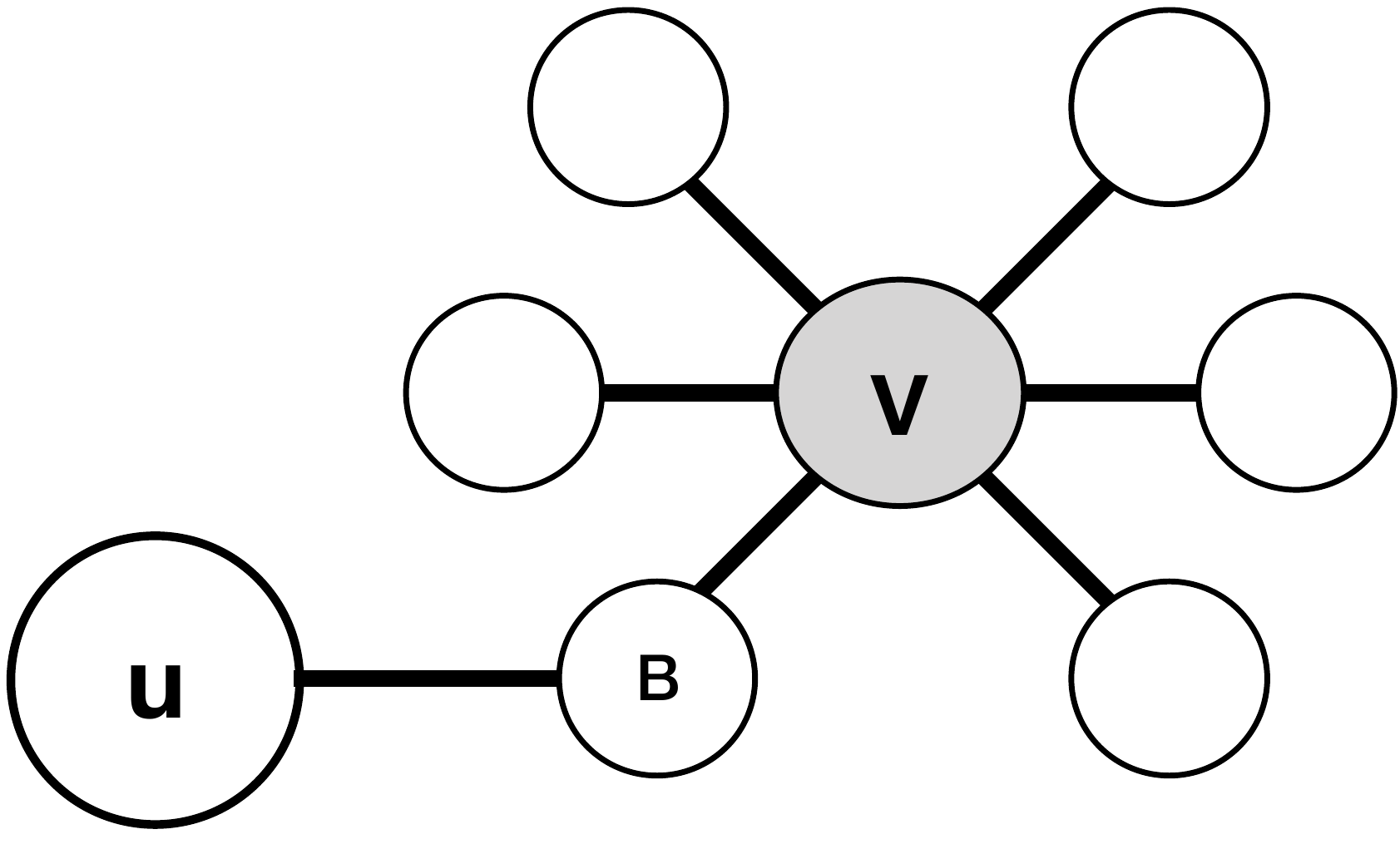}
\caption{\small{Asymmetric local structures in Undirected Graphs}}
\label{fig:app}
\end{center}
\vspace{-0.5cm}
\end{figure}

\subsection{Differences in Optimization Methods}
\label{sec:opt}
 Optimization methods span from direct matrix factorization, deep autoencoders, negative sampling to neighborhood aggregation methods using convolutions.
 
\mpara{Hierarchical softmax \cite{morin2005hierarchical} and Negative Sampling~\cite{mikolov2013distributed}.}
\deepwalk, \nodetovec{}, \lineemb{}-2, \app{} model $f$ in Equation~\eqref{eq:loss} as logarithm of  probability for pair $(i,j)$ sharing an edge in the context graph, i.e., 
\begin{equation}
\label{eq:skip-gramobj}
f(\Phi_i ,\theta_j) = \log{\exp(\Phi_i \cdot \theta_j) \over \sum_k \exp(\Phi_i \cdot \theta_k)}
\end{equation}
\verseemb{} uses exactly the same form of $f$ except that it uses only source representation, i.e, it defines $f$ as
\begin{equation}
\label{eq:verse}
f^{verse}(\Phi_i,\Phi_j) = \log{\exp(\Phi_i \cdot \Phi_j) \over \sum_k \exp(\Phi_i \cdot \Phi_k)}
\end{equation}
Since exact computation of $f$ would require computations over all vertex-pairs which would be very expensive. Instead these methods make use of approximations namely hierarchical softmax and negative sampling. Hierarchical softmax is only used by \deepwalk{}. Other methods employ negative sampling.

For \lineemb-1, the corresponding function is given as 
\begin{equation}
\label{eq:line1}
f^{LINE1}(\Phi_i,\Phi_j) = \log{1 \over 1+\exp(-\Phi_i \cdot \Phi_j) }
\end{equation}
and it further approximates it using negative sampling.

\mpara{Neighborhood Aggregation and Negative Sampling.} \graphsage{} trains a set of aggregator functions that learn to aggregate feature information from a vertex' local neighborhood. Like other methods, \graphsage{} uses an unsupervised loss and its context graph corresponding to the loss function is same as that of \deepwalk. Instead of directly learning the embeddings as done by other methods,  \graphsage{} learns the parameters of the aggregator functions via stochastic gradient descent. 

\mpara{Deep Autoencoders.} \sdne ~uses a multi-layer auto-encoder model to capture non-linear structures based on first- and second-order proximities. By reconstructing first order proximity, the model aims to embed vertices closer which have links between them with the corresponding loss function given by $ \mathcal{L}_1 = \sum_{i,j} c_{i,j} ||\Phi_i -\Phi_j ||^2$.

Drawing parallel to \eqref{eq:loss2} we have $f^{SDNE} = -||\Phi_i -\Phi_j ||^2 $.
For preserving second order proximity it uses the adjacency matrix as input to the autoencoder. Denoting row $i$ of matrix $C$ by $c_{i}$ the reconstruction process
will make the vertices with similar neighborhood structures have similar latent representations, i.e, the following loss function will be minimized: $\mathcal{L}_2 = \sum_{i} ||c_i - g(\Phi_i) \odot \beta||^2$.
where $g$ is a decoder function. $ \mathcal{L}_2 $ is an auxiliary reconstruction loss and is restricted to a node rather than a pair of nodes and hence is of different form than Equation \eqref{eq:loss2}.

The contribution of these two proximities is controlled by the hyperparameter $\alpha$ such that setting $\alpha=0$ will switch to only preserving second order proximity. Another hyperparameter $\beta$ controls the reconstruction of zero elements in the adjacency matrix of the training graph. For simplicity we state the loss function without the regularization term:
\begin{equation}
\label{eq:sdne}
    \mathcal{J} =   \mathcal{L}_2 + \alpha \mathcal{L}_1 
\end{equation}

\begin{remark}
\label{rem:sdne}
Like \sdne,  \lineemb~also aims to preserve first and second order proximities. But unlike \lineemb, \sdne~uses a deep neural network and performs joint optimization as opposed to learning two separate embeddings and later concatenating them. 
\end{remark}

\mpara{Matrix Factorization.}
\hope{} and \netmf{} compute low rank decomposition of their respective context matrices. While \hope{} uses both factors for downstream task denoting the first factor as the source representation of the vertex and the second as target representation, \netmf{} only uses one representation matrix for downstream tasks. Their loss function is given as: $\mathcal{J} =   || C - \Phi\cdot \theta||_F^2$,
where $||.||_F$ denotes the Forbenius norm.

Table~\ref{tab:algos-setup} summarizes the list of embedding methods along with the corresponding properties with respect to defining and exploiting context and loss functions.

\subsection{ Research Questions }
\label{sec:rq}
Based on the differences due to context and optimization methods, we formulate the following research questions.
\vspace{0.1cm}
\begin{RQ}
\label{rq:edgeweights} How does the choice of different context schemes defined in Section \ref{sec:contextschemes} affect the performance of downstream tasks? And to what extent is this performance influenced by the structural properties of the underlying graph?
\end{RQ}

\begin{RQ}
\label{rq:edgedirectionality} How do different ways of exploiting the context listed in Section \ref{sec:contextexploitation}, affect the performance of network representation learning methods? Which combination of downstream tasks and input graphs could benefit from the explicit use of context embeddings?
\end{RQ}
\begin{RQ}
\label{rq:optimization} How does the choice of optimization method (listed in Section \ref{sec:opt}) affect the performance? Do deep models always outperform the shallow models? 
\end{RQ}

We answer these research questions in Section \ref{sec:results}~based on the observations from extensive experimental comparison and summarize the answers in Section \ref{sec:rqanswers}.

\section{Task Description}
\label{sec:tasks}

In this section we describe the two most popular tasks used for empirically comparing various UNRL methods -- Link Prediction (LP) and Node Classification (NC). We also discuss the shortcomings of previous works with respect to these tasks and propose new experimental settings to overcome the same. We also considered Graph Reconstruction (GR) and Graph Clustering (GC), please refer to the supplementary material (Sections 4 and 5) for details.

\subsection{Link Prediction (LP)} The aim of the link prediction task is to predict missing edges given a network with a fraction of removed edges. In the literature there have been slightly different yet similar experimental settings. A fraction of edges is removed randomly to serve as the \emph{test split} while the residual network can be utilized for training. The test split is balanced with negative edges sampled from random vertex pairs that have no edges between them. While removing edges randomly, we make sure that no vertex is isolated, otherwise the representations corresponding to these vertices can not be learned. 

For directed graphs in addition to the existence of an edge it is also desirable to learn about the directionality of the edge. Therefore, for directed graphs we inverse a fraction of positive edges in the test split in order to create negative edges. For example given an edge $(a,b)$ in the test split we check if $(b,a)$ is also an edge. If not, we replace another negative edge with $(b,a)$ in the negative edge list of the test split. It is trivial to note that methods using only the source representation would not be able to simultaneously predict the existence of edge $(a,b)$ and non existence of edge $(b,a)$.

Tables \ref{tab:link-pred} and \ref{tab:directed-test} present the ROC-AUC (Area Under the Receiver Operating Characteristic Curve) scores for undirected and directed graphs respectively. For each method, the inner product of representation of the pair of vertices normalized by the sigmoid function is employed as the similarity/link-probability measurement.

\begin{remark}
We remark that most of the previous works are lacking in the sense that they only evaluate if the method predicts a link and ignore the edge directionality for directed graphs hence giving an unfair treatment to methods designed specifically for directed graphs like \hope{} and \app{}.
\end{remark}
\begin{remark}
\label{rem:lp}
Note that the difficulty of link prediction in directed graphs will be influenced by its \emph{reciprocity}. 
\end{remark}

\subsection{Multilabel Node Classification (NC)}
Given a graph, each node has one or more labels. We report the Micro-F1 and Macro-F1 scores after a $5$-fold multi-label classification using one-vs-rest logistic regression. The main motivation behind using embeddings for this task is the assumption that the local vertex neighborhood dictates its labels. For example, a republican would have more republican than democrat friends. We use 5 undirected and 3 directed networks for this task. The three directed networks with labels are the citation networks wherein an edge represents a citation relationship.

\mpara{New Baseline.} In order to better judge the difficulty of predicting labels for a particular graph we propose an improved na{\"i}ve baseline, which we call, \maxvote.  In this approach, in order to assign a label to a vertex, only the labels of its immediate neighbors from the training set are considered. In \maxvote, we first split the datasets into training and test set (80-20) and the labels are assigned for the vertices in the test set using only the labels of the neighbors which are part of the training set. For a given node with $k$ labels in the ground truth, we assign it the most frequent $k$ labels of its labelled immediate neighbors.
If less than $k$ neighboring nodes are labelled or the vertex' neighbors  have less than $k$ labels, remaining labels are chosen randomly from the list of all possible labels in the graph. The pseudo-code for the subroutine to label a vertex is shown in Algorithm \ref{alg:maxvote} where $\ell$ denotes the total number of label classes.
\begin{remark}
\label{rm:homophily}
By \emph{homophily} in node classification we understand that similar nodes share the same label. Our baseline method \maxvote{} quantifies \emph{homophily} when similarity is limited to similarity between immediate neighbors. 
\end{remark}

\begin{algorithm} [t!!]
\begin{algorithmic}[1]

\caption{Subroutine to label a node with \maxvote}
\label{alg:maxvote}
    \Function{Max-Vote}{$v, N(v),k$}
\For{$(i=0,1,\ldots \ell)  $} 
                \State{$L(i)=0$}
                \EndFor
            \For{$(i\in N(v))$}
                \If{($i$ is labelled)}
                   \For{$j\in labels(i)$}
                   \State{ $L(j)= L(j)+1$}
                   \EndFor
                    \EndIf
            \EndFor
            \State{Choose the most frequent $k$ labels in $L$  to label $v$}
		 \EndFunction
\end{algorithmic}
\end{algorithm}
\section{Structural Properties}
\label{sec:structprop}
In order to quantify the impact that different kinds of graphs have on the performance of the vertex representations, we consider diameter, reciprocity, clustering coefficient, transitivity and spectral separation. 

In order to compute \textbf{diameter ($D$)}, edge directions are not considered. In networks that are not connected, the diameter of the largest connected component is reported. In a directed network, the \textbf{reciprocity ($r$)} equals the proportion of edges for which an edge in the opposite direction exists, i.e., that are reciprocated. More formally, $r= \frac 1 m  | \{ (u,v) \in E \mid (v,u) \in E \} | $. 

The \textbf{local clustering coefficient} of a vertex quantifies how probable it is for $v$ to form a clique of size $3$ with its neighbors. Formally, if $d(v)$ is the degree of $v$, then local clustering coefficient of $v$ is defined as 
 $$ c(u) = { |(u,w)\in E \mid  (u,v) \in E, (v,w) \in E  | \over {d(v) \choose 2}}   .$$
For directed graphs, the local clustering coefficient of a vertex  $u$  equals the proportion of directed 2-paths starting from $u$ that are completed by a third edge oriented in the same direction as the 2-path.
The clustering coefficient ($clus$) of graph $G$ is then defined as the average of the local clustering coefficients of its vertices. We denote the directed clustering coefficient by $clus_{dir}$.

\textbf{Transitivity ($T$)} measures the extent to which two nodes are related in a network that are connected by an edge. It is defined as the ratio of the number of vertex triplets forming a triangle to the total number of triads (subgraphs of $3$ vertices).  For directed graphs, the transitivity ($T_{dir}$) equals the proportion of directed 2-paths that are completed by a third edge oriented in the same direction as the 2-path. 

The \textbf{spectral separation ($S$)} is the largest absolute eigenvalue of the adjacency matrix divided by the second largest absolute eigenvalue. Low values (slightly larger than one) indicate many independent substructures in the network.

\section{Experimental Setup}
\label{sec:setup}

We empirically validate the impact of various differences among the 9 embedding methods (cf. Table~\ref{tab:algos-setup}) on task performance. For reproducibility we used the authors' implementations whenever available and performed hyperparameter tuning whenever applicable. We provide a detailed description of parameter settings, hardware and software setup in the supplementary material (Section 2). We consider  six social network graphs, four citation  networks and an authorship network with their structural properties summarized in Table \ref{tbl:datasets}. We consider two tasks LP and NC defined in Section \ref{sec:tasks}. We also consider Graph Reconstruction and Graph Clustering tasks. However, due to lack of space their results are discussed in the supplementary material (Sections 4 and 5).

\begin{table*}[t!!!]
	\tabcolsep 5pt
	
	\centering
	\scalebox{1}{
		\begin{minipage}{\linewidth}
			\centering
			\begin{tabular}{c|cccccccccccr}
				\hline
				\bf Category &\bf Dataset & Type & \bf $|V|$ & \bf $|E|$ &$|\mathcal{L}|$& $r$& $D$ & $clus$ & $T$ & $clus_{dir}$ & $T_{dir}$ & $S$\\\hline
				\multirow{5}{*}{\bf Social} & \blog~\cite{tang2009relational}& undir. & 10K & 333K   & 39 & n.a &   5 
    &0.463  & 0.0914 & n.a&n.a &2.18 \\
 & \flickrl~\cite{tang2009relational}~\label{fn:Flickr} & undir. & 80K & 5.90M & 195&n.a &   6
& 0.165 &0.1875  & n.a&n.a & 2.06\\ 

& \youtube~\cite{mislove-2007-socialnetworks}  & undir. & 1.13M & 2.99M &47  & n.a &   21 & 0.080 &0.0062  & n.a&n.a & 1.19\\
     & \reddit~\cite{hamilton2017inductive} & undir. & 231K & 11.6M &41 &n.a &   10 & 0.169 &  0.0458 & n.a&n.a& 1.47 \\
&\twitter~\cite{de2010does}& dir. & 465K & 834K  &n.a & 0.3\% &   8 & 0.0006 & 0. 0152 &0.0002 & 0.013   &1.05 \\
 & \epi~\cite{richardson2003trust} & dir. & 75K  & 508K &n.a& 40.52\%	 &   15 & 0.1378 & 0.0657   &0.0982 &0.0902  & 1.74 \\\hline

                & \dblpc~\cite{ley2002dblp}& dir. & 12.5K & 49K &n.a & 46.4\% &   10  & 0.1169 & 0.0620  &0.039 &0.0967 & 1.39\\
				
               \multirow{3}{*}{\bf Citation} & \cocit~\cite{sinha2015overview} & dir. & 44K& 195K &15 & 0\% &   25 &0.1419&0.0806&0.0826  & 0.0913 & 1.07	 \\
                	& \cora~\cite{vsubelj2013model}~\label{fn:cora} & dir. & 23K & 91K &70 & 5.00\% &   20 &  0.2660 &0.1169 & 0.169 & 0.221 & 1.03 \\
                    & \pubmed~\cite{namata2012query} \label{fn:pubmed} & dir. & 19K & 44k &3 &0.07\% &  18 & 0.0602 & 0.0537 & 0.0325 &0.0530  & 1.14
\\\hline
\multirow{1}{*}{\bf Collaboration}
                 & \dblpa~\cite{tang2015line}& undir. & 1.2M & 10.3M   & n.a&n.a &   24 &0.635 & 0.1718 & n.a&n.a& 1.0005\\\hline

\end{tabular}
	\end{minipage}}
	\caption{\small{A summary of benchmark datasets with properties : number of nodes ($|V|$), number of edges ($|E|$), number of labels ($|\mathcal{L}|$), reciprocity ($r$), diameter ($D$), clustering coefficient ($clus$), transitivity ($T$), directed clustering coefficient ($clus_{dir})$ and transitivity ($T_{dir}$), spectral separation ($S$). 'n.a' indicates the specific property is 'not applicable' to the corresponding graph. }}
	\label{tbl:datasets}
\end{table*}

With respect to datasets, \blog, \flickrl{} and \youtube{} are social networks with users as nodes and friendship between them as undirected edges with multiple labels per node. \twitter~ and \epi~ are unlabelled, directed graphs modeling the  follower and trust between users respectively. \dblpc, \cocit, \cora, \pubmed~are directed graphs representing academic citation networks, with vertices as papers and edges representing the citations between them. \dblpa~is a collaboration network of authors of scientific papers from DBLP Computer Science bibliography. An elaborate description about the datasets can be found in the Supplementary material.

\section{Results and Discussion}
\label{sec:results}

\begin{table}[ht!!]
\begin{center}
\renewcommand\arraystretch{1.35}
\begin{tabularx}{\textwidth}{p{1.5cm}ccccc}
\multicolumn{1}{l}{\emph{method}} & BlogCat. & \youtube & \reddit & \dblpa   &  \flickrl\\
\rule[0.005cm]{240pt}{0.5pt}\\

\small\dw & 0.527  & 0.586 & 0.897 & 0.850  & 0.772\\

\rowcolor{lightgray}
\small\nodetovec{}  & 0.556  & 0.652 & 0.892 & 0.949 &  0.821\\

\small\verseemb  & \textbf{0.878} & 0.884  & 0.973 & \textbf{0.994}  & 0.918\\
\rowcolor{lightgray}
 \small\app{}  & 0.790 & 0.871 & \textbf{0.974} & \textbf{0.994}  & \textbf{0.928} \\

\small NetMF  & 0.659 & \xmark  & 0.949 & \xmark  & 0.604\\
	\rowcolor{lightgray}
\small\lineemb-{1+2}  & 0.612  & \textbf{0.894} & 0.949 & 0.989   & 0.839 \\

\small\lineemb-{1}  & 0.495 & 0.758 & 0.947 & 0.989  & 0.830 \\
\rowcolor{lightgray}
\small\lineemb-{2}  & 0.400 & 0.823 & 0.833 & 0.896  & 0.694 \\

\small GraphSage  &0.619 & 0.778  & 0.936 & 0.912  & 0.734\\
\rowcolor{lightgray}
\small  GSage(GCN)  &0.661 & 0.813  & 0.941 & 0.975  & 0.779\\

\small SDNE  & 0.519 & \xmark & \xmark & \xmark  & 0.483\\
\rule[0.35cm]{240pt}{0.75pt}\\
\end{tabularx}
\caption{\small{Link prediction results for undirected graphs using 50\% edges as training data. \xmark~indicates the corresponding method failed to finish for the given dataset.}}
\label{tab:link-pred}
\end{center}
\end{table}

\label{sec:graph-obj}

\subsection{Link Prediction}
\label{sec:lp}
Main results for the LP task for both undirected (cf. Table \ref{tab:link-pred}) and directed graphs (cf. Table~\ref{tab:directed-test}) are summarized below:
\begin{enumerate}
\item For undirected graphs, PPR based methods -- \app{} and \verseemb{} -- are more or less the best performing methods in all datasets (cf. Table \ref{tab:link-pred}).

\item \lineemb{} that directly uses adjacency matrix as context matrix outperforms random walk based methods for undirected graphs (cf. Section \ref{sec:lpcontext}).
\item For directed graphs with low reciprocity, context representation of a node plays a major role (cf. \ref{sec:lpcontext}) and methods encoding and using two embedding spaces for source and target roles of nodes should be used for directed link prediction. 
\item Deeper models do not have a considerable advantage over the shallow ones for this task 
(cf. Section \ref{sec:diffopt}).
\end{enumerate}

\subsubsection{Different Schemes of Context.} 
\label{sec:lpcontext}

In this section, we investigate in detail the performance difference potentially caused by differences in the definition of the context as questioned in RQ\ref{rq:edgeweights}. 

\mpara{Random Walk Based Approaches.} We first question the utility of computationally-expensive biased walks employed by \nodetovec{} and establish that the biased walks do in fact perform worse than simpler counterparts like \deepwalk~for graphs with certain structural properties.
 On the contrary, for undirected graphs with high clustering ratio like \blog{}, one observes a relatively higher standard deviation (computed mean and standard deviations provided in Supplementary Material) from the mean of scores computed with 25 combinations of the $p$ and $q$ parameters. Similarly, for directed graphs with high reciprocity and high clustering coefficient, the choice of parameters $p$ and $q$ matters for \nodetovec.
Note that from properties \ref{prop:param1node2vec} and \ref{prop:param2node2vec}, the biased walks of \nodetovec{} will not produce any significant gains for graphs with low \emph{clustering coefficient} and \emph{low reciprocity} for example \twitter{}. This is also evident in the empirical results (see Table \ref{tab:directed-test}).
 Notable differences are only observed for directed dataset with high reciprocity and clustering coefficient, i.e., \epinion{} where \nodetovec{} outperforms \deepwalk{} by $72.86\%$ for the case when only random negative edges exist in the test set. 
We also observe that for other directed graphs with high reciprocity and clustering coefficient, \nodetovec{} performs better than \deepwalk{}.
Hence we infer that the biased walks in \nodetovec{} can produce considerably different results from \deepwalk{} for graphs with high clustering coefficient, high diameter and high reciprocity (in case of directed graphs).

\begin{table*}[ht!!]
\begin{center}
\setlength{\tabcolsep}{1pt}
\renewcommand{\aboverulesep}{0pt}
\renewcommand{\belowrulesep}{0pt}
\renewcommand\arraystretch{1.35}
\newcolumntype{C}{>{\centering\arraybackslash}X}

\begin{tabularx}{\textwidth}{p{4cm}CCC|CCC|CCC|CCC}
\multicolumn{1}{c}{} & \multicolumn{3}{c}{\cora}& \multicolumn{3}{c}{\twitter}&\multicolumn{3}{c}{\dblpc}&\multicolumn{3}{c}{\epi}\\
\cmidrule{2-4}  \cmidrule{5-7}  \cmidrule{8-10}\cmidrule{11-13}
\multicolumn{1}{l}{\emph{method}} & 0\% & 50\% & 100\% & 0\% & 50\% & 100\% & 0\% & 50\% & 100\% & 0\% & 50\% & 100\% \\
\hline
\small\dw & 0.836 & 0.669 & 0.532 & 0.536 & 0.522 & 0.501 & 0.868 & 0.680& 0.503 & 0.538 & 0.560 & 0.563 \\
\rowcolor{lightgray}
\small\nodetovec{}  &0.840 & 0.649 & 0.526 &  0.500&  0.500 & 0.500 & 0.889 & 0.697&0.503 & 0.930 & 0.750 & 0.726 \\
\small\verseemb{}  & 0.875 & 0.688 & 0.500 &0.52& 0.510 & 0.501 & 0.809 & 0.654 & 0.503 & \textbf{0.955} & \textbf{0.753} & \textbf{0.739} \\
\rowcolor{lightgray}

\small\app{}  & 0.865 & \textbf{0.841} & \bf{0.833} & 0.723& 0.638 & 0.555 & \textbf{0.957} & \textbf{0.838}& 0.722 & 0.639 & 0.477 & 0.455 \\
\rowcolor{lightgray}
\small \hope  & 0.784 & 0.734 & 0.718 & \textbf{0.981}& \textbf{0.980} & \textbf{0.979} & 0.756 & 0.737& \textbf{0.732} & 0.807 & 0.718 & 0.716\\

\small\lineemb-{1+2}  & 0.735 & 0.619 & 0.518 & 0.009 & 0.255 & 0.500 & 0.319 & 0.404 & 0.501 & 0.658 & 0.622 & 0.617 \\
\rowcolor{lightgray}
\small\lineemb-{1} & 0.781 & 0.644 & 0.526  & 0.007 &  0.007 &  0.254 & 0.312 & 0.405 & 0.501 & 0.744 & 0.677 & 0.668 \\	
\small\lineemb-{2}  & 0.693 & 0.598 & 0.514 & 0.511 & 0.507 & 0.503 & 0.642 & 0.572 & 0.503 & 0.555 & 0.544 & 0.543 \\

\rowcolor{lightgray}
\small \graphsage  &0.902 & 0.707 & 0.531 & 0.659& 0.602 & 0.504 & 0.806 & 0.656 & 0.503 & 0.814 & 0.672 & 0.658 \\
\small \graphsagegcn  &\bf{0.927} & 0.721 & 0.534 & 0.589& 0.539 & 0.502 & 0.856 & 0.670 & 0.503 & 0.816 & 0.668 & 0.668 \\
\rowcolor{lightgray}

\small \sdne  & 0.613 & 0.557 & 0.507 & \xmark & \xmark & \xmark & 0.569 & 0.540 & 0.501 &  0.601 & 0.560 & 0.551\\

\bottomrule
\end{tabularx}
\caption{\small{Link Prediction Results for directed graphs with (1) random negative edges in test set (2) 50\% of the test negative edges created by reversing true edges of the test set (3) when all true edges of test set are reversed to create negative edges in the test set. \xmark~indicates the corresponding method failed to finish for the given dataset. }}
\label{tab:directed-test}
\end{center}
\end{table*} 

\begin{table*}[t]
\begin{center}
\setlength{\tabcolsep}{1pt}
\renewcommand{\aboverulesep}{0pt}
\renewcommand{\belowrulesep}{0pt}
\renewcommand\arraystretch{1.35}

\newcolumntype{C}{>{\centering\arraybackslash}X}

\begin{tabularx}{\textwidth}{p{4cm}|CC|CC|CC|CC|CC|CC|CC}
\multicolumn{1}{p{3cm}}{} & \multicolumn{2}{c}{\blog}& \multicolumn{2}{c}{\pubmed} & \multicolumn{2}{c}{\cora} & \multicolumn{2}{c}{\reddit} & \multicolumn{2}{c}{\flickrl} & \multicolumn{2}{c}{ \youtube}& \multicolumn{2}{c}{\cocit}\\
\cmidrule{2-15}
\multicolumn{1}{l}{\emph{method}} & mic. & mac. & mic. & mac. & mic. & mac. & mic. & mac. & mic. & mac. & mic. & mac. & mic. & mac.\\
\midrule

\small\dw{} & 42.15 & 28.48 & 73.96&71.34 & 64.98&51.53 &94.40&92.01 & \textbf{42.20} & \textbf{31.00}& 47.09&39.89& 41.92&30.07 \\

\rowcolor{lightgray}
\small\nodetovec{}  &42.46&\bf{29.16}& 72.36&68.54& 65.74&49.12 & 94.11&91.73& 	42.11&30.57&\bf{48.41}&\bf{42.04} & 41.64&28.18 \\

\small\verseemb{} &  35.51&21.77 & 71.24&68.68  & 60.87&45.52& 92.87&89.69&  35.70 &	23.00 &45.12&37.28 & 40.17 & 27.56 \\
\rowcolor{lightgray}
\small\app{} & 20.60&5.39& 69.00&65.20 & 64.58&47.03&  77.11&56.28 & 24.26&4.21 & 45.04&36.61 & 40.34&28.06 \\
\rowcolor{lightgray}
\small \hope  & n.a & n.a & 63.00 & 54.6 & 26.23 & 1.22  & n.a & n.a & n.a & n.a & n.a & n.a & 16.66 & 1.91 \\
\small\netmf&\textbf{43.29}&29.04 &73.66&71.11 & 63.38&46.16 & 91.99&86.92&37.44&21.55 & \xmark & \xmark &40.42&28.7\\
	\rowcolor{lightgray}
\small\lineemb-{1+2}  & 41.01&25.02 &  62.29&59.79& 54.04&41.83 &\textbf{94.50}&\bf{92.08} & 41.46&27.65 &48.22&41.51&37.71 & 26.75  \\
\small\lineemb-{1}  & 41.54&24.28 &  55.65&53.83& 62.36&47.19 &94.31&91.96 & 40.92&26.19 & 47.49&41.17&36.10&25.70  \\
\rowcolor{lightgray}
\small\lineemb-{2}  & 36.70&18.80 &  56.81&51.71& 51.05&35.37&94.30&91.81 & 40.49&24.24 & 47.46&39.97&31.4&20.59  \\
\small \graphsage &19.28&5.07 &77.90&76.39 &67.07&44.78 & 89.94& 82.28 &25.52&5.84 & 40.45&29.97 & 43.71&30.52 \\
\rowcolor{lightgray}
\small \graphsagegcn &26.76&10.82 & \textbf{79.19}&\textbf{77.85} &69.64&51.64 & 91.65&86.88 &29.66&9.69 & 42.54&32.54& 44.08&30.73 \\
\small SDNE & 26.40&12.29 & 46.41 & 32.32  & 32.43&8.27 & \xmark & \xmark  & 29.10 & 10.53  & \xmark & \xmark  &21.67&9.53 \\

\rowcolor{lightgray}
\small \maxvote  & 32.71& 19.60 & 76.81 & 75.25  & \bf{71.96} & \bf{57.21} & 93.26 & 90.11 & 34.60 & 22.48 & 28.96 & 25.65  & \bf{44.66}&\bf{33.39}\\

\bottomrule
\end{tabularx}
\end{center}
\caption{\small{Multilabel Node Classification results in terms of Micro-F1 and Macro-F1. All results are mean of 5-fold cross validations. \xmark~indicates the corresponding method failed to finish for the given dataset. 'n.a' indicates the given method is 'not applicable' to the corresponding graph.} }\label{tab:node-classification}
\end{table*} 

\mpara{Adjacency based approaches.} We observe that for link prediction on undirected graphs, \lineemb~performs better than \deepwalk~and \nodetovec. 
Note that \lineemb~uses adjacency matrix as its context graph. 
We observe that for \youtube{}, that has the lowest clustering co-efficient and transitivity, \lineemb~outperforms \deepwalk{} and \nodetovec~by $26.99\%$ whereas for \flickrl~with transitivity of $0.1875$, the gain is $3.1\%$. 
On the other hand, for \dblpa~and \flickrl~with high clustering coefficient and transitivity, \lineemb~outperforms these methods by a smaller margin. 
All of the observations lead to the conclusion that \lineemb{} performs comparable or better compared to \deepwalk~and \nodetovec{}, with the performance becoming much better for graphs with low clustering coefficient and transitivity. 
SDNE on the other hand performs worse than \lineemb{} and other methods (for more discussion see Section \ref{sec:diffopt}). 

\mpara{Direct Matrix based approaches.} \netmf{} is designed specifically for undirected graphs and \hope{} for directed graphs. 
In the original paper \netmf{} was not compared for the task of link prediction. \netmf{} could only be run for smaller graphs and there is no clear advantage of using \netmf{} over other methods for link prediction task. 
Of the three datasets we observe that \netmf{} performs better than \deepwalk{},
\nodetovec{} and \lineemb{} for  \blog{} and \reddit{} with low transitivity but high clustering coefficient. 
\hope{} while using two embedding spaces to encode a vertex in its source and target roles outperforms most of the single embedding based methods for directed link prediction but is still mostly outperformed by \app{}, exceptions being for \twitter~and \epi. Interestingly, \hope{} is better in predicting the edge direction than \app{} (see results corresponding to 100\% edge reversal in Table~\ref{tab:directed-test}). 
In summary, Katz-based context graph as used by \hope{} performs best for directed graphs with very low reciprocity and low clustering coefficient for example \twitter.

\mpara{\graphsage{}.} \graphsagegcn~outperforms \deepwalk, \nodetovec{} and \sdne{} for most of the directed and undirected graphs. It's performance is  comparable to \lineemb{} for undirected graphs with high transitivity.
For the directed graph, \cora{} which has high transitivity it outperforms not only \lineemb{} but also \hope{} and \app{} when only random negative edges are considered in the test set (see the $0\%$ columns for \cora{} in Table \ref{tab:directed-test}). 
To understand the attributing reason, we note that in graphs with high transitivity there would be many cases such that for nodes $u,v,z$ : $(u,v)\in E, (v,z)\in E \implies (u,z)\in E$. 
Let edge $(u,z)$ be in the test set and the other edges are in the training set. 
In such a case we would expect our embedding method to treat nodes $u$ and $z$ similar (embed them closer) because of existing edges $(u,v)$ and $(v,z)$. Recall that in a GCN architecture, in each layer the hidden representations are averaged among neighbors that are one-hop away. Therefore after 2 layers, a node representation is some aggregation function of its neighbors which are one and two hops away.
Now we recall that in two layered neighborhood aggregation based methods like \graphsagegcn{}, a representation of a node is an aggregation of representations of its one-hop neighborhood which in turn is an aggregation of its neighborhood, thereby encoding similarity between two-hop neighborhoods.  In other words the aggregation steps smoothens the node representation along the two hop edges in the graph, thereby encouraging similar representations for two-hop neighbors. We remark that the differences between \graphsage{} and \graphsagegcn{} are because of the aggregation operators and we empirically observe that for most of the datasets convolution based aggregator performs better.

\subsubsection{Exploitation of Context }
\label{sec:lpcontext}
We recall that both methods \app~and \verseemb~use similar context graphs, but the main difference is that \verseemb{} uses a single embedding space, while \app~uses two different embedding spaces. The advantage of using two embedding spaces by \app{} for undirected link prediction is not clear. As per the authors, differing local properties of nodes such as degree, may cause asymmetries in undirected graphs. This argument is still insufficient to interpret the use of the embedding space to predict missing links. For example, assume we want to predict the existence of a link between the nodes $u$ and $v$. Using two embedding spaces might predict a link between $u$ and $v$ but not between $v$ and $u$. This can happen when the destination representation of $v$ is embedded closer to the source embedding of $u$ but the source embedding of $v$ is far away from the destination $u$. Note that such a result is probable for example for vertices $u$ and $v$ as shown in Fig. \ref{fig:app}. For undirected LP task, other methods (such as \lineemb{}-1 and \graphsagegcn{}) learning a single representation and hence explicitly preserving symmetric properties of source and context  perform relatively better than \deepwalk{}, \nodetovec{} and \lineemb{}-2 which learn both source and context representations.

\mpara{Effect of Reciprocity in Directed Link Prediction.} For directed graphs with low reciprocity, learning and using two embedding matrices per vertex is more intuitive as these two matrices represent the two roles of a vertex (source and target respectively). We observe that single embedding based methods are insufficient to capture the directed relationship in graphs. We report results corresponding to the LP for directed graphs in Table \ref{tab:directed-test}. Note that we test three settings: for $0\%$, we use random negative edges in the test set, for $50\%$ and $100\%$ we force the model to not only predict the right edges but also decide on the edge direction by using the reverse of true (positive) edges in the test set as negative edges (if possible). Methods using two representations per vertex, \hope~and \app~outperform single embedding based methods for all directed datasets except for \epi~which has a high reciprocity.

\subsubsection{Differences in Optimization.}
\label{sec:diffopt}
We observe that the joint optimization of first and second order objectives using deep auto-encoder by \sdne{} does not provide any additional performance gains as compared to \lineemb{} and other methods. We could not run \sdne{} for bigger datasets because of its prohibitive memory requirements imposed by the input adjacency matrix.

\graphsage{} shares the same unsupervised loss function as \deepwalk{} but instead of learning directly embeddings it learns parameters of neighborhood aggregation functions. Even though it is not the best performing method when compared to its counterpart sharing the same loss function, it performs much better than \deepwalk{} for link prediction in directed and undirected graphs.

\subsection{Node Classification}
\label{sec:homophily}
 In this section, we look at the results from node classification (Table~\ref{tab:node-classification}). We also present additional experiments to measure the learning rate of different methods for the NC task in supplementary material (Section 4). In contrast to the earlier link prediction task, node classification is a supervised task that includes external information in the form of labels. The effectiveness of an unsupervised representation for vertex classification is the extent to which it can reconcile varying degrees of homophily. We observe that \deepwalk~is either the best performing approach or reasonably competitive in most of the datasets. Note that this is both \emph{surprising and counter intuitive} since it was the earliest proposed approach. This calls into question the utility of its other variants for example biased random walk methods such as \nodetovec.

As mentioned in Remark~\ref{rm:homophily} the homophily baseline (\maxvote) measures the degree of label similarity among neighboring nodes. Lower values indicate low neighborhood homophily where a node is less likely to share the label of its neighbours. In Section~\ref{sec:ed} we investigate the effect of neighbourhood homophily in detail and its consequence on utility of edge direction in directed graphs.

 \subsubsection{Different Schemes of Context.}
 \mpara{Random Walk based.} Both \deepwalk{} and \nodetovec{} perform quite well in this task. Taking advantage of longer walks exploiting similarities with higher order neighborhoods, both of these methods perform specifically well when \maxvote's performance's drops, i.e., when neighboring nodes might not have the same labels and it is required to consider labels of higher order neighborhoods.

\mpara{PPR based.} In contrast to link  prediction, for node classification task PPR based context matrix is not the best performing. 

\mpara{Direct Matrix  based.} We observe that \netmf~has the best Macro-F1 for NC task on \blog~and close to \deepwalk~in all other cases. We could not run \netmf~for large datasets because of prohibitive memory requirements limiting any further analysis. \hope{} performs poorly for all datasets. We believe that as \hope{} is tied to a particular similarity matrix, it is limited to a certain type of task and cannot be generalized.

\begin{table*}[ht!!!]
		\tabcolsep 5pt

		\centering

			\begin{tabular}{|c|c|c|c|}
				\hline
				  \bf Algorithm &\bf Favourable Task & \bf \tabincell{c}{Favourable Label Properties} & \bf Favourable Graph Properties   \\
				\hline
			\dw & NC & \tabincell{c}{Robust for  different label distributions}  & High Spectral separation \\
		\hline
				
			 \nodetovec & NC  & \tabincell{c}{Robust for  different label distributions}  & High Clustering Coefficient, High Reciprocity   \\
			\hline
			
         \app & LP  & - & Directed Graphs, Low Spectral Separation\\
        \hline
          \verseemb  & LP  & -  & Undirected or directed with high reciprocity   \\
        \hline
         \lineemb& LP, NC  & \tabincell{c}{Low Similarity \\ among labels of neighboring nodes} & Undirected, low clustering coefficient, low transitivity\\
                           \hline
          \netmf  & NC  &Robust for different label distributions & Undirected \\
          \hline
          \hope & LP  & - & \tabincell{c}{Directed Graphs with Low Reciprocity \\ and low Clustering coefficient} \\
 \hline
 \graphsagegcn & LP, NC & \tabincell{c}{High Label  similarity\\ among immediate neighbors}& Undirected graphs with high clustering coefficient  \\
\hline
		\end{tabular}
		\caption{\small{Summary of Main Results corresponding to best performing methods.}
}
		\label{tab:results}
	\end{table*}
\subsubsection{Exploitation of Context}
\label{sec:ed}
We believe that for directed graphs, edge directionality has little effect on the labels on the nodes. 
We verify this hypothesis through the empirical performance of various methods as shown in Table~\ref{tab:node-classification} and as discussed below. Qualitatively we can argue that an edge in the studied directed graphs represents a citation relationship between two papers. Now each paper cites and gets cited by papers of similar areas of labels, hence limiting the effect of direction of citation. 

\mpara{Baseline - \maxvote } First, we observe that our na{\"i}ve \maxvote~baseline, which ignores the edge direction, outperforms other methods for all directed graphs except PubMed.  

\mpara{\app{} and \hope{}.} For all directed graphs, methods ignoring the context representation outperform \hope~and \app~which  use both vertex and context representation for node classification. 

\mpara{\verseemb{} and \lineemb{}-1.} \verseemb{} which only learns vertex representation, hence ignoring the role of context, performs better than \app{} which shares the same context graph as \verseemb{} but additionally learns and uses context representations.
Moreover, \lineemb-1 which specifically only learns vertex representation, hence ignoring the edge directionality, outperforms \lineemb-2 (designed to take directionality into account) in almost all datasets (except \pubmed). 

\mpara{\deepwalk{}, \nodetovec{}, \lineemb{}-2, \netmf{}.} As already observed in Property~\ref{prop:symm} the context matrix used by these methods is symmetric even if the underlying graph is directed. Consider for example, a directed random walk of length $1$ with vertex sequence $(v1,v2)$ with window length $1$. Now the following source-context pairs are considered for training: $(v1,v2)$ and $(v2,v1)$, thereby ignoring the edge direction between $v1$ and $v2$. Note that this is not the case for \lineemb-2 which would only consider $(v1,v2)$ for training. So in the above described sense, \deepwalk~and \nodetovec~are still ignoring edge directionality, even if they operate on directed random walks.

 \subsubsection{Differences in Optimization}
 Similar to link prediction, \lineemb{} outperforms \sdne{} which uses deep autoencoders. \graphsagegcn{} outperforms most of the other methods when the  baseline \maxvote{} performs well, i.e., when there is a high degree of similarity among neighboring nodes. This is also understandable since \graphsagegcn{} constructs node representations as convolutions of its immediate neighbor representations which explains its good performance when there exists a strong homophily in label distribution of neighboring nodes.

\subsection{Answers to Research Questions}
\label{sec:rqanswers}
In the following we summarize the above analysis while answering the research questions from Section~\ref{sec:rq}. The main results pertaining to favourable tasks and dataset properties for best performing methods are summarized in Table~\ref{tab:results}.

\mpara{RQ \ref{rq:edgeweights}.} From the experimental results described above, it is clear that the choice of different schemes of defining context graph is dependent on both task and underlying graph properties. For example, for the LP task, PPR based context graph construction provides best results for both undirected and directed graphs. Similarly, for directed graphs, with low reciprocity, Katz similarity based context graph provides best results. Biased walks on the other hand have advantages in link prediction for graphs with high clustering coefficient, high transitivity and high reciprocity. With low clustering coefficients and transitivity \lineemb-1 performs much better for link prediction. Random walk based methods are more robust in the task of node classification while neighborhood  aggregation  based  methods  perform  best  if  there is a high similarity among labels of neighboring nodes. 
    
   \mpara {RQ \ref{rq:edgedirectionality}.} Context representations should be explicitly used in directed link prediction. The lower the reciprocity (i.e. higher assymetricity in the role of vertex as source and context) in directed graphs, the more  important are the context representations. For undirected graphs, methods learning only single node representations (hence preserving the symmetric nature of links) outperform others. 
   Explicit modelling of edge direction via context in directed graphs does not provide any advantage in NC task for the widely used (in other papers) citation networks.

    \mpara{RQ \ref{rq:optimization}.} In general, single layer models using negative sampling work better for both LP and NC tasks. Neighborhood aggregation learnt via SGNS objective works best for node classification when there is a high similarity among labels among neighboring nodes. Optimizing first and second order proximities using negative sampling based objective as done by \lineemb{} is better than using deep autoencoders to encode these proximities as done by \sdne{}. 
    
Finally, we also provide best practices and caveats for the practitioners in the supplementary material (Section 4).

\section{Conclusions}
 We  studied 
the important but unexplored problem of analyzing  differences between widely used network representation learning approaches. To the best of our knowledge, we are the first to compare UNRL methods using (1) a common unifying framework based on the concept of context, (2) structural properties of the underlying graph and (3) large-scale experiments to demonstrate the properties of various methods observed by the common framework. Our analysis provided several non-intuitive insights which are beneficial for practitioners and academics to apply network embedding techniques for graphs with different properties and different tasks.

\section{Acknowledgements}
This work is partially funded by SoBigData (European Union’s Horizon 2020 research and innovation programme under
grant agreement No. 654024).

\bibliographystyle{abbrv}
\bibliography{references} 
	\vspace*{-3\baselineskip}
\begin{IEEEbiography}[{\includegraphics[width=1in,height=1.25in,clip,keepaspectratio]{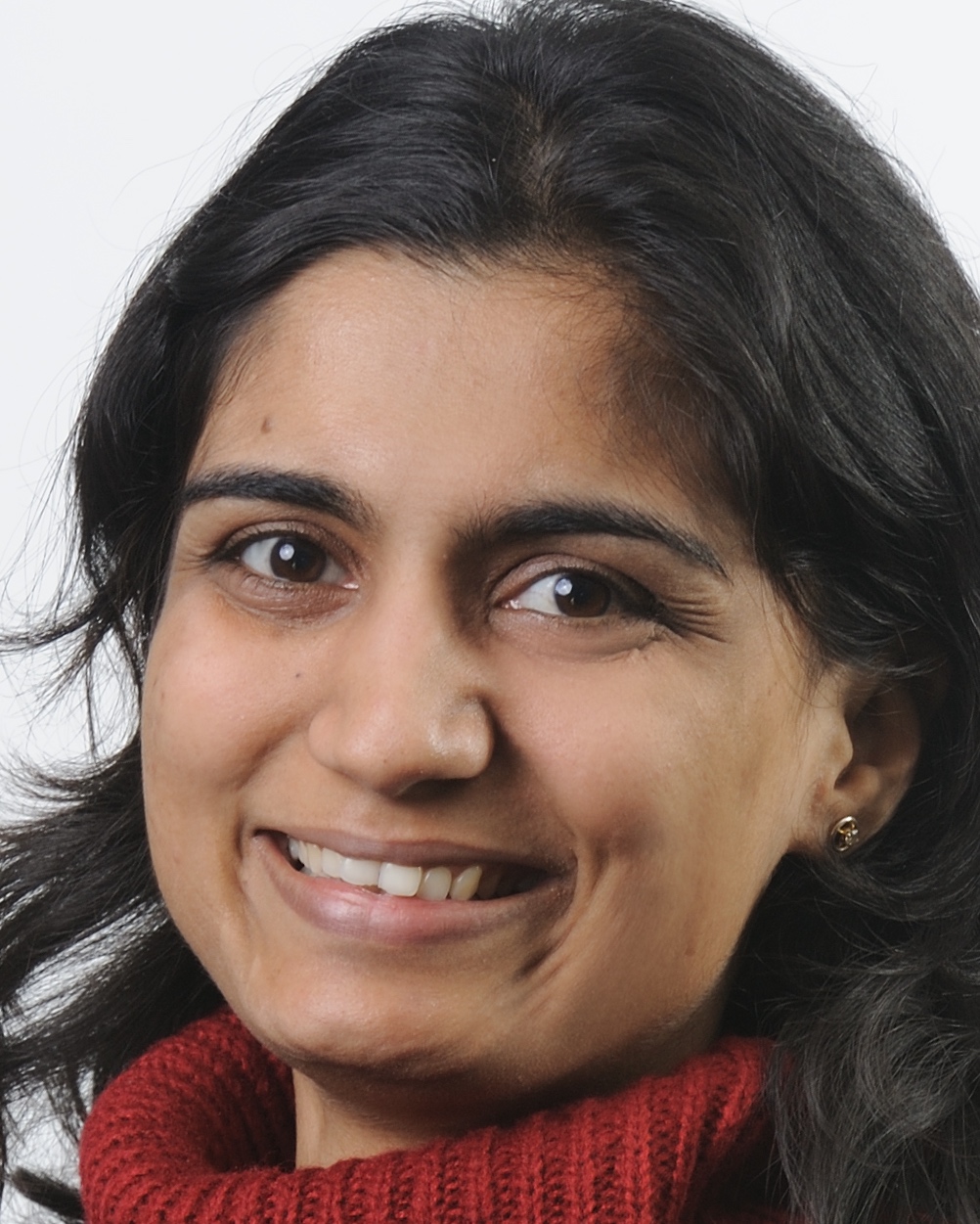}}]{Megha Khosla} received her PhD degree in Theoretical Computer Science from Max Planck Institute of Informatics and Saarland University, Saarbruecken, Germany. Currently she is a post doctoral researcher at L3S Research center. Her main research focus is on Mining and Learning on Large Graphs.
\end{IEEEbiography}
\vspace*{-3\baselineskip}

	\begin{IEEEbiography}[{\includegraphics[width=1in,height=1.25in,clip,keepaspectratio]{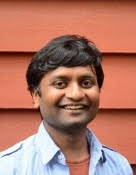}}]{Vinay Setty}
		is an Associate Professor at the Department Electrical Engineering and Computer Science, University of Stavanger, Norway. In addition to graph mining and network embeddings, his research areas include news ranking, mining and classification.
	\end{IEEEbiography}
	\vspace*{-3\baselineskip}
\begin{IEEEbiography}[{\includegraphics[width=1in,height=1.25in,clip,keepaspectratio]{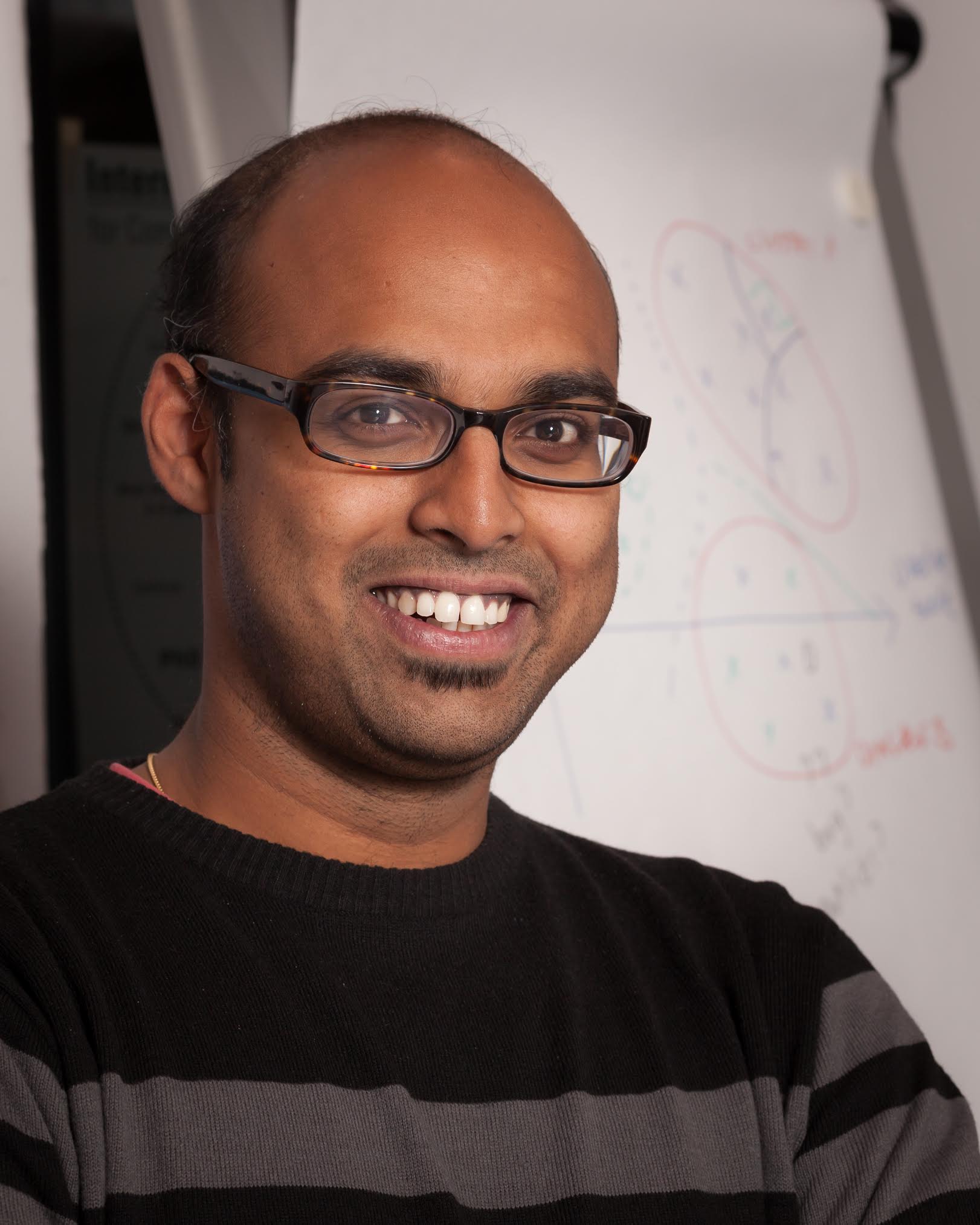}}]{Avishek Anand} is an assistant professor in the Department of Knowledge Based Systems, Leibniz Universität Hannover, and a member of the L3S Research Center, Hannover, Germany. One of his main research focus is scalable learning of continuous representations from discrete input like text and graphs.
\end{IEEEbiography}



\section*{Supplementary material}

# A Comparative Study for Unsupervised Network Representation Learning (Suplementary Material)

Megha Khosla[1], Vinay Setty [2], and Avishek Anand[1]

[1]L3S Research Center and Leibniz Universität Hannover, Germany
[2]Department of Electrical Engineering and Computer Science, University of Stavanger, Norway

October 30, 2019

## 1 Missing Proofs

*Proof of Proposition 1.* Note that $i$ is chosen uniformly at random , i.e., the probability of choosing $i$ as the starting vertex is $\frac{1}{|V|}$. For an $h$ hop walk starting from fixed $i$, the probability that $j$ is the chosen neighbor is given by the $(i,j)th$ element of the transition matrix over $h$-hops, i.e., $\mathbf{P}^h_{i,j}$.

Again the probability that the walk will stop in exactly $h$ hops is $(1-\alpha)^h \cdot \alpha$. Again there might exist paths of length greater than $h$ from $i$ to $j$; say of lengths $h_1, h_2 \ldots h_k$. Then probability that pair $(i,j)$ will be sampled is given by

$$\Pr((i,j) \text{ is sampled}) = \frac{1}{|V|} \sum_{h' \geq h} (1-\alpha)^{h'} \cdot \alpha \cdot (\mathbf{P}^{h'})_{i,j} \quad (1)$$

$$\leq \frac{1}{|V|} \sum_{h' \geq h} (1-\alpha)^{h'} \cdot \alpha \cdot (\mathbf{P}^h)_{i,j}$$

$$= \frac{1}{|V|} (1-\alpha)^h \cdot \alpha \cdot (\mathbf{P}^h)_{i,j} \sum_{h' \geq 1} (1 + (1-\alpha)^{h'})$$

$$\leq \frac{2}{|V|} (1-\alpha)^h \cdot \alpha \cdot (\mathbf{P}^h)_{i,j}$$

□

## 2 More details on Datasets

**Social Network Graphs:** BlogCatalog, Flickr and Youtube are social networks with users as nodes and friendship between them as undirected edges. All these datasets also have multiple labels per node for each group or community the user belongs to. Reddit is an artificially generated network by [4], connecting Reddit posts if the same user comments on both. Each node has a single label representing the corresponding reddit community. Twitter and Epinion are unlabelled, directed graphs modeling the follower and trust between users respectively. The labels for the nodes in Reddit graph represent the subreddit (communities) they belong to. Since each post can only belong to one subreddit, each node has only one label.

**Citation Graphs:** DBLP-Ci, CoCit, Cora, PubMed are directed graphs representing academic citation networks, with vertices as papers and edges representing the citations between them. While DBLP-Ci is unlabelled, Cora has multilabels representing the sub-communities in Computer Science such as "Machine Learning", "Databases" etc. DBLP-Ci and Cora are parsed from scientific publications from the Computer

Table 1: NODE2VEC parameters for link prediction task

| Dataset | p | q | mean | stddev |
|---|---|---|---|---|
| BlogCatalog | 4 | 4 | 0.543 | 0.0048 |
| Flickr | 2 | 4 | 0.811 | 0.0031 |
| Reddit | 4 | 0.25 | 0.889 | 0.0027 |
| DBLP-Au | 0.25 | 4 | 0.947 | 0.0008 |
| Youtube | 0.25 | 4 | 0.615 | 0.0167 |
| Cora | 4 | 4 | 0.838 | 0.00083 |
| Twitter | 0.25 | 4 | 0.500 | 0.0000 |
| DBLP-Ci | 0.5 | 4 | 0.800 | 0.0343 |
| Epinion | 0.25 | 4 | 0.888 | 0.0233 |

Science community. While DBLP-Ci is unlabelled, Cora has multilabels representing the sub-communities in Computer Science. CoCit is a labeled citation graph from Microsoft Academic Graph, with labels representing conferences in which the papers were published. Finally, PubMed, is a citation graph derived from the medical literature database pertaining to diabetes classified into one of three classes of diabetes.

**Collaboration Network:** DBLP-Au is a collaboration network of authors of scientific papers from DBLP Computer Science bibliography. An undirected edge between two authors represents a common publication. There may be multiple edges between the authors if they collaborated on multiple papers, but we only consider single edges.

## 3 Parameter Settings

Here we describe all the tunable hyperparameters which are common across methods. Unless specified explicitly we use default parameters provided by the author implementations. For all methods we fix the embedding dimensions $d = 128$ as it is the most common practice in the literature.

**Random Walk:** For all methods which rely on random walks, we set the target walk length $t = 40$ and number of walks $r = 80$ as it provided the best results. For all methods using SGNS, we set the negative sample size $ns = 5$ and window size $w = 10$. For all methods we also set number of worker threads to 32 since we observed a minor variation in performance with different number of worker threads. This is due to the way random walks are performed in parallel.

For NODE2VEC there are two hyperparameters $p$ and $q$, for biased random walks. The authors recommend exploring the parameters $p, q \in \{0.25, 0.5, 1, 2, 4\}$. Since that results in 25 combinations, it is very expensive to explore these parameters for all datasets especially for large datasets such as Youtube and Reddit. For these datasets we fix the $p = 0.25, q = 4$ which were the best performing parameters in most cases. We summarize the best performing parameters in Table 1 and 2 along with mean and standard deviation of the accuracy values for different values of $p$ and $q$. As you can observe, these parameters do not play a huge role in the performance of NODE2VEC as the standard deviation is quite low in most cases.

For VERSE, we fix $\alpha = 0.85$, which is the default setting used for personalized page rank algorithm in [11]. We omit the variation HVERSE which is nothing but the best performing accuracies after hyperparameter exploration in the original paper since it is too expensive.

For APP, no information is provided in the original paper about the optimal parameters, therefore we iterate through the node list 80 times, in each iteration we run 10 random walks per node, thereby totalling 800 random walks per node as we do with all the random walk based methods.

For LINE, we run experiments with $T = 10$ billion samples and $s = 5$ negative samples, as described by the authors in their paper [10]. In addition, we also compare three variants of LINE: (1) LINE-1 (LINE with first-order proximity), (2) LINE-2 (LINE with second-order proximity) and (3) LINE-1+2 which is obtained by normalizing and concatenating the 64-dimensional embedding vectors from LINE-1 and LINE-2.

**Matrix Factorization:** For HOPE, we set the attenuation factor $\beta = 0.01$ for all datasets except PubMed for all tasks. For PubMed, best results were obtained at $\beta = 0.5$. Choosing optimal $\beta$ is difficult

Table 2: NODE2VEC parameters for node classification task

| Dataset | $p$ | $q$ | Mean mic.F1 | Stddev mic.F1 | Mean mac.F1 | Stddev mac.F1 |
|---|---|---|---|---|---|---|
| BlogCatalog | 0.25 | 4 | 41.87 | 0.587 | 28.44 | 0.610 |
| PubMed | 0.25 | 0.25 | 72.01 | 0.230 | 68.03 | 0.238 |
| Cora | 0.25 | 4 | 65.30 | 0.268 | 47.66 | 0.737 |
| Reddit | 0.25 | 4 | - | - | - | - |
| Flickr | 0.25 | 2 | 41.57 | 0.990 | 29.53 | 1.971 |
| Youtube | 0.25 | 4 | - | - | - | - |
| CoCit | 0.5 | 0.25 | 41.56 | 0.059 | 28.05 | 0.102 |

as only a rough guideline is available, i.e., $\beta$ should be less than 1 divided by spectral radius of adjacency matrix to ensure the convergence of Katz measure. The authors reported best results for Cora at $\beta = 0.1$. We therefore searched for best value of $\beta$ lying close 0.1 and reported the best results.

For NETMF, we set number of eigenpairs (rank) $h = 256$ for BlogCatalog and $h = 16384$ for Flickr as suggested by the authors in their paper [8]. For rest of the datasets we set the default value of $h = 256$. We also observed that setting negative sample value $ns = 5$ as with other random walk approaches resulted in significantly worse performance in some cases. Therefore, we resorted to the default value of $ns = 1$. In addition, for NetMF, the authors provide two different ways to compute Eigenvector decomposition, by specifying the parameters –small and –large which corresponds to small and large window lengths respectively. For smaller datasets we tried both and report the best performing numbers but for large datasets such as Flickr and Reddit, NETMF could only finish with –small. For many larger datasets such as Youtube and DBLP-Au, NETMF crashed by exhausting main memory before finishing training. Since, NETMF requires symmetric adjacency matrix as input, for node classification task, we convert the directed graphs to undirected and create a symmetric matrix. However, for link prediction, such a conversion does not make sense since we consider the directionality of the edges for link prediction task.

**Deep Learning:** For deep learning methods we consider GRAPHSAGE and SDNE. Since the authors do not provide any implementation for SDNE, we use a public implementation in keras. For BlogCatalog we use the hyperparameters such as hidden layer size recommended by the authors in their paper [12]. However, for Flickr the recommended later configuration resulted in "ResourcesExhausted" error. Furthermore, we explored $\alpha = 50, 100$ and $\beta = 1, 5, 10$ parameters for SDNE with Flickr dataset without any significant improvements.

For GRAPHSAGE, since we only deal with transductive, unsupervised setting in this paper we only use the unsupervised version. GRAPHSAGE provides four aggregators: Mean, MeanPool, MaxPool and LSTM. We repeat all experiments with each aggregator and report the best values. We also include a variant of GRAPHSAGE with GCN aggregator (GraphSAGE-GCN). Since it is significantly different from other GRAPHSAGE aggregators, we report it separately. There are several hyperparameters such as learning rate, dropout, epochs, batch size etc. It is extremely expensive to tune all these parameters for all the datasets. Instead, we follow the recommendations of the authors and explore the learning rate in 0.001, 0.0001, 0.00002 [4, 12]. For GRAPHSAGE, in [9], the authors perform a grid search over several of these hyperparameters and they recommend "Mean" aggregator, along with learning rate of 0.0001, dropout 0.4 for inductive setting. The authors also recommend using "–model_size big" option for unsupervised setting which we follow. The results of GRAPHSAGE could be further improved by performing more exhaustive exploration of hyperparameters. However, we do not expect any contradictions to our findings.

# 4 Graph Reconstruction

In the graph reconstruction task we evaluate how well the embeddings preserve neighborhood information of the original graph. In our experiments, we use the evaluation measure proposed in [5] which is more general

and is applicable for undirected and directed graphs as it not only measures the performance of embeddings on reconstructing the outgoing edges of a node but also the incoming edges (which makes a difference for directed graphs). The other two evaluation schemes as used in [7] and [11] suffer from several drawbacks as highlighted in [5]. We summarize here the schemes and their issues her for completeness. The *edge-centric* evaluation in [7] relies on sampling random pairs of nodes from the original graphs into their test set. These candidate edges are then ordered according to their similarity in the embedding space. Precision is computed at different rank depths where the relevant edges are the ones present in the original graph. On the other hand, [11] perform a *node-centric* evaluation where precision is computed on a per-node basis. For a given node $v$ with an outdegree $k$, embeddings are used to perform a $k$-nearest neighbor search for $v$ and precision is computed based on how many actual neighbors the $k$-NN procedure is able to extract.

The edge-centric evaluation suffers from sparsity issues typical in real-world networks and even if a large number of node pairs are sampled, the fraction of relevant edges retrieved tends to remain low. More acutely, such an approach does not model the neighborhood reconstruction aspect of graph construction and is rather close to predicting links. The adopted measure from [5] is closer to node-centric evaluation approach where we intend to also compute precision on directed networks with a slight modification.

**Evaluation Measure.** In particular, we compute precision for both outgoing and incoming edges for a given node. This is different from teh other two evaluation approaches which only considers the reconstruction of adjacency list of a node, i.e., only its outgoing neighbors. Moreover in this strategy the prior knowledge of the indegree or outdegree is not assumed.

As in Link Prediction, the similarity or the probability of an edge $(i, j)$ is computed as the sigmoid over the dot product of their respective embedding vectors. For HOPE and APP we use the corresponding source and target vectors respectively. We do not assume the prior knowledge of the indegree or outdegree, rather we compute the precision for $k \in \{1, 2, 5, 10, 100, 200, 500, 1000\}$. For a given $k$ we obtain the $k$-nearest neighbors ranked by sigmoid similarity for each embedding approach. If a node has an outdegree or indegree of zero, we set the precision to be 1 if the sigmoid corresponding to the nearest neighbor is less than 0.51 (recall that $\sigma(\vec{x} \cdot \vec{y}) = 0.5$ for $\vec{x} \cdot \vec{y} = 0$), otherwise we set it to 0. In other cases, for a given node $v$ and a specific $k$ we compute $P_{out}^k(v)$ and $P_{in}^k(v)$ corresponding to the outgoing and incoming edges as

$$P_{out}^k(v) = \frac{\mathcal{N}_{out}^k \cap N^{out}(v)}{k}, \quad P_{in}^k(v) = \frac{\mathcal{N}_{in}^k \cap N^{in}(v)}{k},$$

where $\mathcal{N}_{out}^k(v)$ and $\mathcal{N}_{in}^k(v)$ are the $k$ nearest outgoing (to whom $v$ has outgoing edges) and incoming (neighbors point to $v$) neighbors retrieved from the embeddings and $N^{out}(v)$ and $N^{in}(v)$ are the actual outgoing and incoming neighbors of $v$. We then compute the Micro-F1 score as the harmonic mean of $P_{in}^k(v)$ and $P_{out}^k(v)$. To avoid any zeros in the denominator, we add a very small $\varepsilon = 10^{-5}$ to each precision value before computing the harmonic mean. We finally report the final precision as the average of these harmonic means over the nodes in the test set. We conduct experiments for the Graph Reconstruction task with four undirected graphs and three directed graphs. For the experiments, we randomly sample 10% of the nodes and use them as test set. We use the best performing parameters for the node classification task for each method.

**Results.** The results are presented in Figure 1. Unsurprisingly the two best performing methods for the directed graphs are HOPE and APP which use two embedding spaces to encode separately node as a source and as a context. The results indicate that HOPE is a better choice than APP. From the comparison of HOPE and APP we can conclude that HOPE should be a preferred choice when the downstream task involves reconstructing the original graph from the embeddings. This also points to better suitability of using Katz similarity measure as compared to PPR to construct the context graph.

For undirected graphs, on the other hand, APPis the worst performing method. This is because it uses two embedding spaces for encoding a node as a source and context even for undirected graphs and fail to represent the symmetric property of undirected edges. In particular, it might predict existence of an outgoing edge $(u, v)$ but non existence of incoming edge $(v, u)$ which is clearly not true in an undirected graph. Note that as the used evaluation scheme explicitly checks for incoming and outgoing neighbors, we see a drop in APP's performance which might not be so clear if we had only evaluated on predicting the adjacency list of a vertex.

Considering the undirected graphs, NODE2VEC, LINE1+2 and DEEPWALK are the top performing methods. Note that for all 4 undirected networks, we observe performance differences between NODE2VEC and

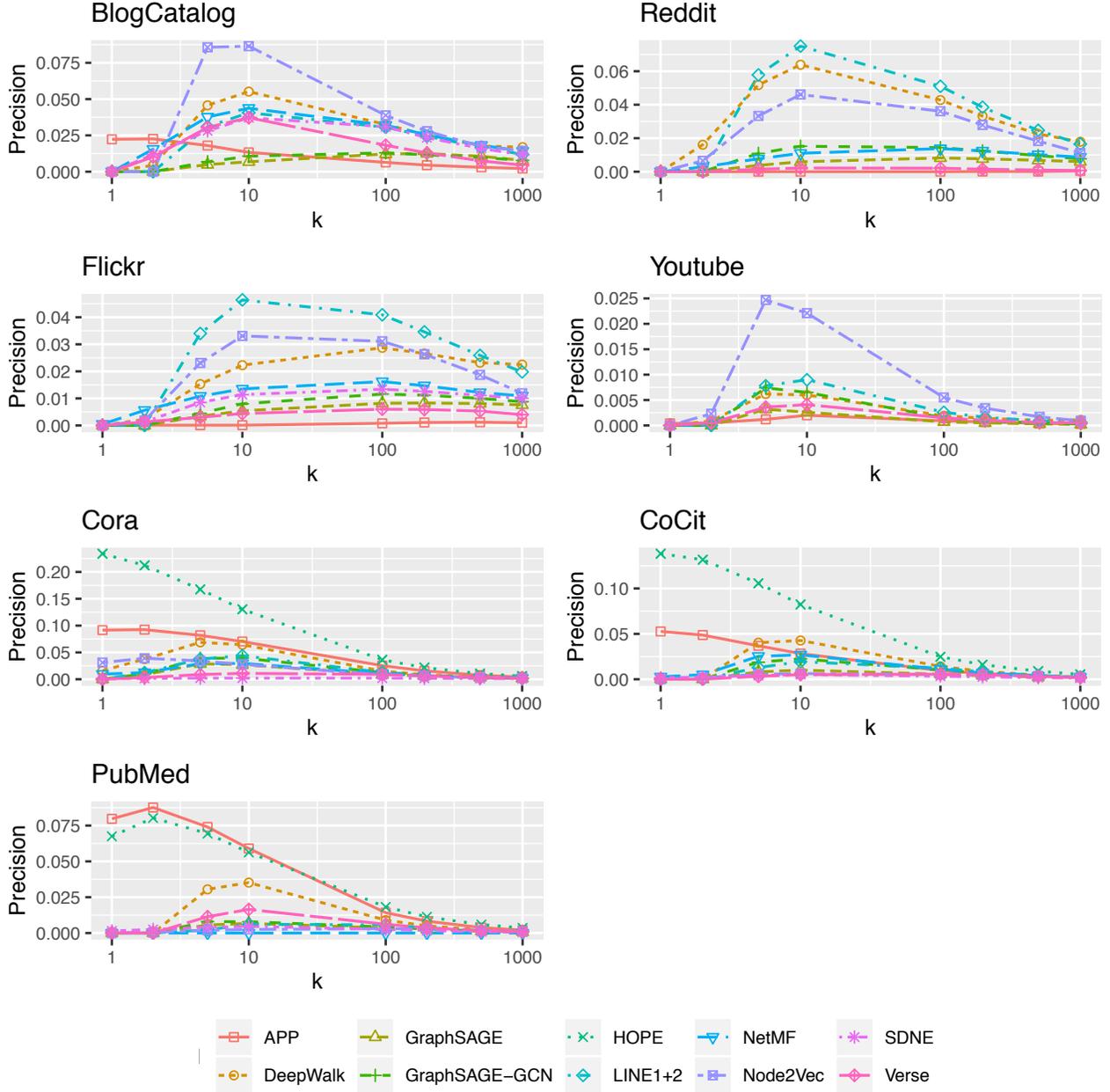

Figure 1: Results for Graph Reconstruction task with randomly chosen 10% nodes as test set. X axis is number of neighbors $k$ (log scale), Y-axis is precision

DEEPWALK. This is because all these graphs have either high clustering coefficient ( for example BlogCatalog, Reddit, Flickr) or high diameter (Youtube). Recall that for cases with high clustering coefficient, transitivity and diameter, the biased walks are capable of sampling much different neighborhoods as compared to DEEPWALKleading to performance differences among the two methods. As in Link Prediction, LINE1+2 performs better than both of these methods for Reddit and Flickr.

## 5 Graph Clustering

Graph clustering task is used to detect groups of nodes with similar characteristics. Following the setup from existing works such as [2], we perform $K$-means clustering with number of clusters varying from 2 to

5

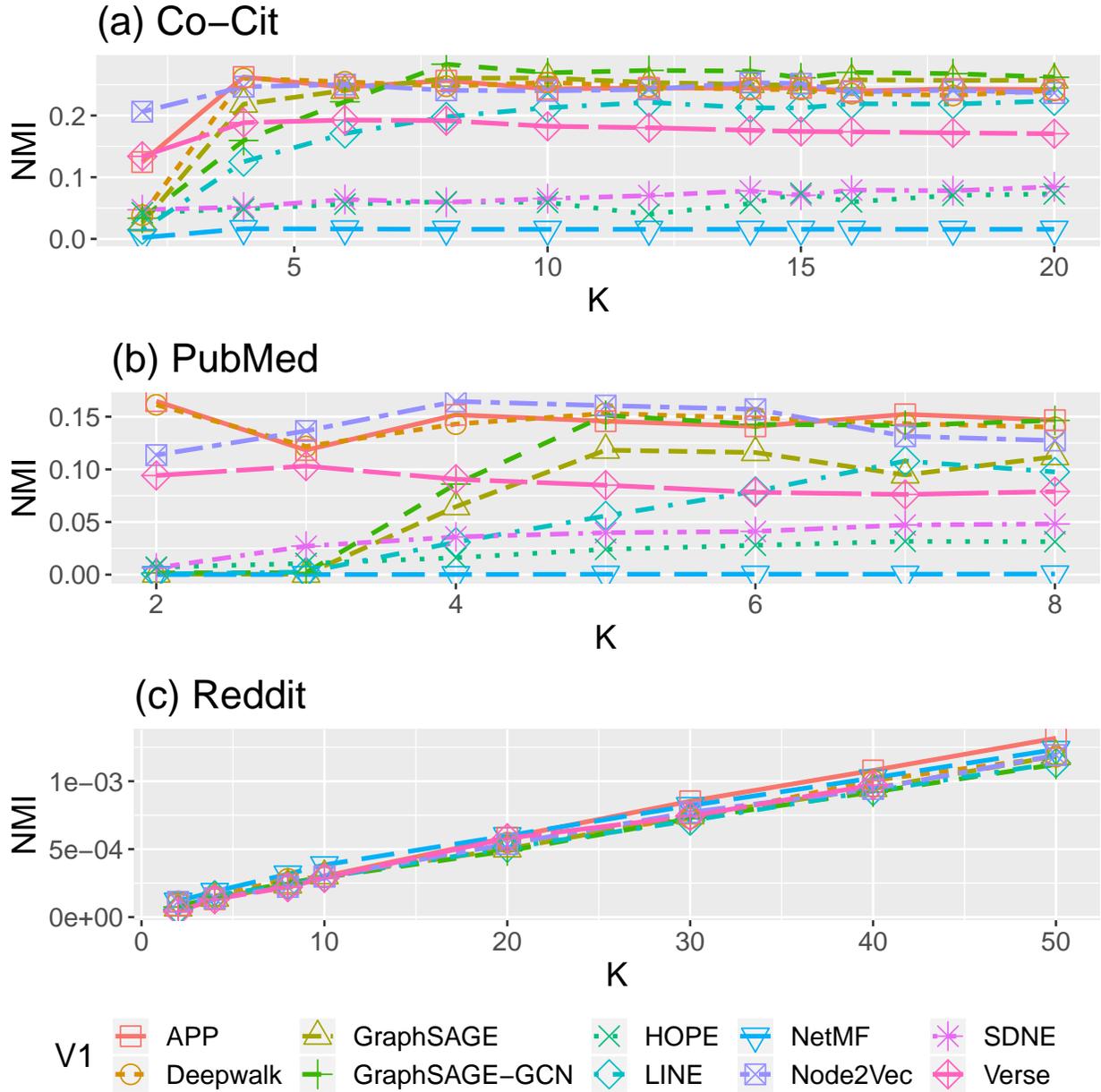

Figure 2: Results for Graph Clustering task. NMI score is plotted corresponding to different number of clusters (K)

number of labels in the graph. This is more natural since, based on the node classification task we expect that nodes with same label cluster together.

In Figure 2, we show the Normalized Mutual Information (NMI) score with respect to the number of clusters ($K$). Since the NMI is computed w.r.t the node labels, which corresponds to the clusters produced by the clustering algorithm, we can only compute it for datasets with single label per node. Most of the datasets have multi-labels per node. That leaves us Reddit, CoCit and PubMed.

**Results.** From Figure 2 it is clear that for both CoCit and PubMed, GraphSAGE-GCN, Node2vec, DeepWalk, APP are all close to each other and perform reasonably well. On the other hand NetMF, HOPE and SDNE are the two worst performing methods. Note that since the graph clustering quality is measured based on the node labels, the results are fairly similar to the node classification task. The

performance of HOPE as is also the case in node classification is the worst.

The reason for this is two-fold: First, as already discussed in the main paper, encoding edge directionality does not give any foreseeable advantage in predicting labels. Second, as also pointed out in [5], HOPE is tied to Katz similarity measure and is not generalizable across tasks.

The performance of various methods for Reddit is quite similar as also is the case in the classification task. This also raises concerns over use of Reddit for evaluating node embeddings as done in [1]. Note that the naive baseline MAX-VOTE already achieves a very high F1 score for Reddit in classification task which points to extremely high homophily and reduced complexity of the dataset as compared to others.

# 6 Hardware and Software

We train all embeddings on Linux servers with 80 core Intel Xeon 2.40GHz CPU, 1TB main memory running "Scientific Linux" distribution. For algorithms which need GPUs we use Nvidia Tesla P100 GPU units with 16GB memory. Most algorithms were executed using Python 2.7 with the exception of APP and HOPE which are implemented in C++ and MatLab respectively. For SDNE, keras 2.4.4 with tensorflow 1.11 backend and GraphSAGE was executed with tensorflow 1.11.

# 7 Discussion and Best Practices

From our experiments we conclude that while using unsupervised methods for downstream tasks such as link prediction and node classification it is important to be cognizant to graph properties, label distribution and certain best practices in the experimental setup. We performed one additional experiment for node classification, common in many papers, for observing the trends in learning improvements when the training data is steadily increased. The practice employed by notable works [3, 11] fix a training data sample and take the complement as the test set. We contend this by instead fixing the test data set (to 20% of the input) as choosing a variable test set is misleading. We increase the training data in steps of 10% (cf. Figure 3). Note that we do not report all the approaches due to legibility of the plot (but they show similar trend).

We observe not surprisingly that the performance increases with increasing training data and plateaus at 40% of the training data with some exceptions. GraphSAGE still continues to learn with increasing training data for Cora and BlogCatalog. On the contrary most of the approaches for CoCit already converge to their final performance values at 20% of training data. This suggests, contrary to small training datasets in earlier works that consider as small as 1% training data size, one should at least consider at least 20% of training data while reporting performance values.

**Threats to validity.** We chose the datasets in a manner that at least one of the datasets is used in the paper for an approach. We further chose to experiment with the authors implementation as much as possible except for APP and SDNE. We also were able to replicate the results mentioned in the original paper except SDNE for NC task on BlogCatalog and Flickr. Finally, we re-trained models as and when necessary and made them stronger using newer datasets or reverted to the best parameters suggested in the original papers. However, we did not explore all hyper-parameters in all approaches due to their sheer combinatorial search space. We report and verify all the structural properties as mentioned in Konect [6] and compute those which are missing.

## 7.1 Advice to practitioners

In employing node embeddings for tasks like node classification and link prediction some of the key aspects to bear in mind in the choice of the approach are the following.

1. When considering an undirected graph for link prediction PPR based methods such as VERSE and APP are recommended.

2. For doing link predictions in directed graphs almost always node and context embedding pairs like APP and HOPE should be preferred. Only in cases when the reciprocity of the graph is high the other approaches become competitive. In terms of evaluation one should carefully construct test sets with negative edges as reversed positive edges to evaluate directionality.

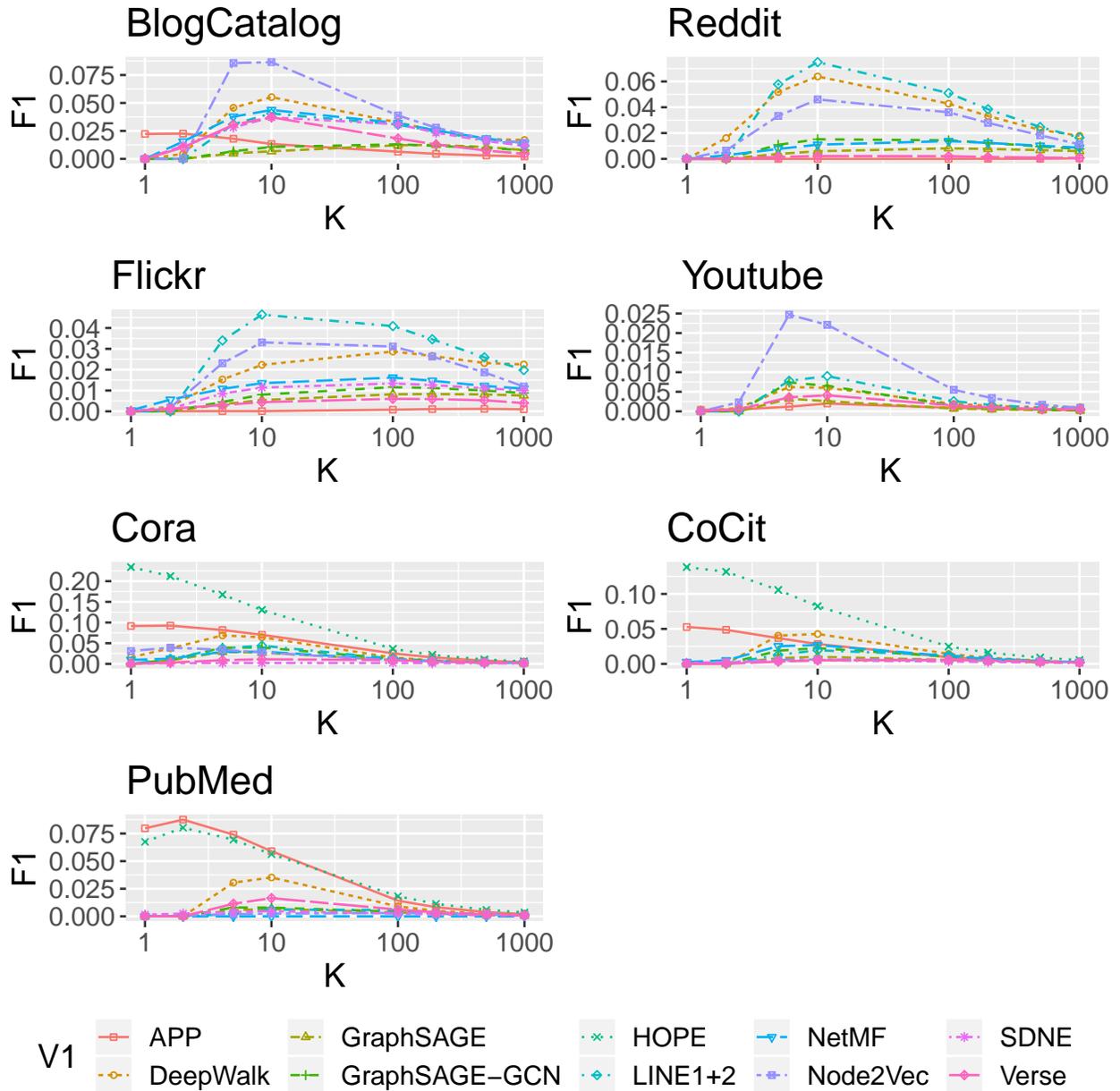

Figure 3: Learning rate with increasing Training data size. All runs are averaged over 5 splits of 20% Test data.

3. For directed graphs, HOPE is recommended over APP when the downstream task is graph reconstruction.

4. For node classification the degree of homophily should be precomputed and that should drive the choice of the method. For high degree of label homophily among neighboring nodes neighborhood aggregation based deep learning approaches outperform others, while DeepWalk is a robust choice for low homophily graphs.

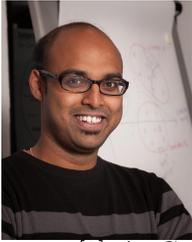

**Avishek Anand** is an assistant professor in the Department of Knowledge Based Systems, Leibniz Universitt Hannover, and a member of the L3S Research Center, Hannover, Germany. One of his main research focus is scalable learning of continuous representations from discrete input like text and graphs.



\end{document}